\documentclass[11pt]{article}

\usepackage[preprint]{acl}

\usepackage{times}
\usepackage{latexsym}

\usepackage[T1]{fontenc}
\usepackage[utf8]{inputenc}

\usepackage{microtype}

\usepackage{inconsolata}

\usepackage{graphicx}

\usepackage{xpinyin}
\usepackage{CJKutf8}
\usepackage{adjustbox}
\usepackage{booktabs} 
\usepackage{graphicx}
\usepackage{subfigure}
\usepackage{multirow}
\usepackage{xcolor}
\usepackage{pifont}
\usepackage{calc}
\usepackage{xurl}
\usepackage{times}
\usepackage{bbding}
\usepackage{hyperref}
\usepackage{siunitx}
\usepackage[most]{tcolorbox}
\usepackage[table,xcdraw]{xcolor}
\usepackage{latexsym}
\usepackage{float}
\usepackage{listings}
\usepackage{array}
\usepackage{multirow}
\usepackage{algorithm}
\usepackage{algorithmic}
\usepackage{amsmath}
\usepackage{graphicx}
\usepackage{booktabs}
\usepackage{amssymb}
\usepackage{makecell}
\usepackage{enumitem}
\usepackage{svg}
\usepackage{longtable}
\usepackage{caption}
\usepackage{multirow}
\usepackage{tabularx}
\usepackage{pifont}% http://ctan.org/pkg/pifont
\usepackage{listings}
\usepackage{array}
\usepackage{booktabs}

\newcolumntype{C}[1]{>{\centering\arraybackslash}m{#1}}
\newcolumntype{L}[1]{>{\raggedright\arraybackslash}p{#1}}

\newtcolorbox{promptbox}[2][]{
  width=\textwidth,            
  boxrule=2pt,
  fontupper=\footnotesize,
  fonttitle=\bfseries\color{black},
  arc=5pt,
  rounded corners,
  colframe=black,
  colbacktitle=white!97!blue,
  colback=white!97!blue,
  title={#2},
  #1
}

\newcommand{\NAME}{ProHist-Bench}
\newcommand{\zh}[1]{\begin{CJK}{UTF8}{gbsn}#1\end{CJK}}

\newcommand{\keju}{\textit{Keju}}
\newcommand{\celun}{\textit{Celun} Generation}

\usepackage{newfloat}
\usepackage{listings}
\DeclareCaptionStyle{ruled}{labelfont=normalfont,labelsep=colon,strut=off} % DO NOT CHANGE THIS
\floatstyle{ruled}
\newfloat{listing}{tb}{lst}{}
\floatname{listing}{Listing}

\title{Can LLMs Act as Historians? Evaluating Historical Research Capabilities of LLMs via the Chinese Imperial Examination}
\author{
 \textbf{Lirong Gao\textsuperscript{1,3}$^*$},
 \textbf{Zeqing Wang\textsuperscript{2,3}$^*$},
 \textbf{Yuyan Cai\textsuperscript{2,3}$^*$},
 \textbf{Jiayi Deng\textsuperscript{3}$^*$},
\\
 \textbf{Yanmei Gu\textsuperscript{3}},
 \textbf{Yiming Zhang\textsuperscript{1}$^\dagger$},
 \textbf{Jia Zhou\textsuperscript{2}$^\dagger$},
 \textbf{Yanfei Zhang\textsuperscript{2}$^\dagger$},
 \textbf{Junbo Zhao\textsuperscript{1,3}$^\dagger$}
\\
 \textsuperscript{1}State Key Laboratory of Blockchain and Data Security, Zhejiang University\\
 \textsuperscript{2}Zhejiang University \quad 
 \textsuperscript{3}Ant Group
\\
\texttt{\{gaolirong,12343006,22443022,yimingz,0012802,yanfei.zhang,j.zhao\}@zju.edu.cn},
 \\
 \texttt{\{dengjiayi.djy,guyanmei.gym\}@antgroup.com}
}

\begin{document}
\maketitle

\footnotetext[1]{Equal contribution}
\footnotetext[2]{Corresponding authors}

\begin{abstract}
While Large Language Models (LLMs) have increasingly assisted in historical tasks such as text processing, their capacity for professional-level historical reasoning remains underexplored. Existing benchmarks primarily assess basic knowledge breadth or lexical understanding, failing to capture the higher-order skills—such as evidentiary reasoning—that are central to historical research. To fill this gap, we introduce \textbf{\NAME}, a novel benchmark anchored in the \textit{Chinese Imperial Examination} (\textit{Keju}) system—a comprehensive microcosm of East Asian political, social, and intellectual history spanning over 1,300 years. Developed through deep interdisciplinary collaboration, \NAME{} features 400 challenging, expert-curated questions across eight dynasties, accompanied by 10,891 fine-grained evaluation rubrics. Through a rigorous evaluation of 18 LLMs, we reveal a significant proficiency gap: even state-of-the-art LLMs struggle with complex historical research questions. We hope \NAME{} will facilitate the development of domain-specific reasoning LLMs, advance computational historical research, and further uncover the untapped potential of LLMs. We release \NAME{} at \url{https://github.com/inclusionAI/ABench/tree/main/ProHist-Bench}.

\end{abstract}

\section{Introduction}
\begin{figure}[t]
    \centering
    \includegraphics[width=1.0\linewidth]{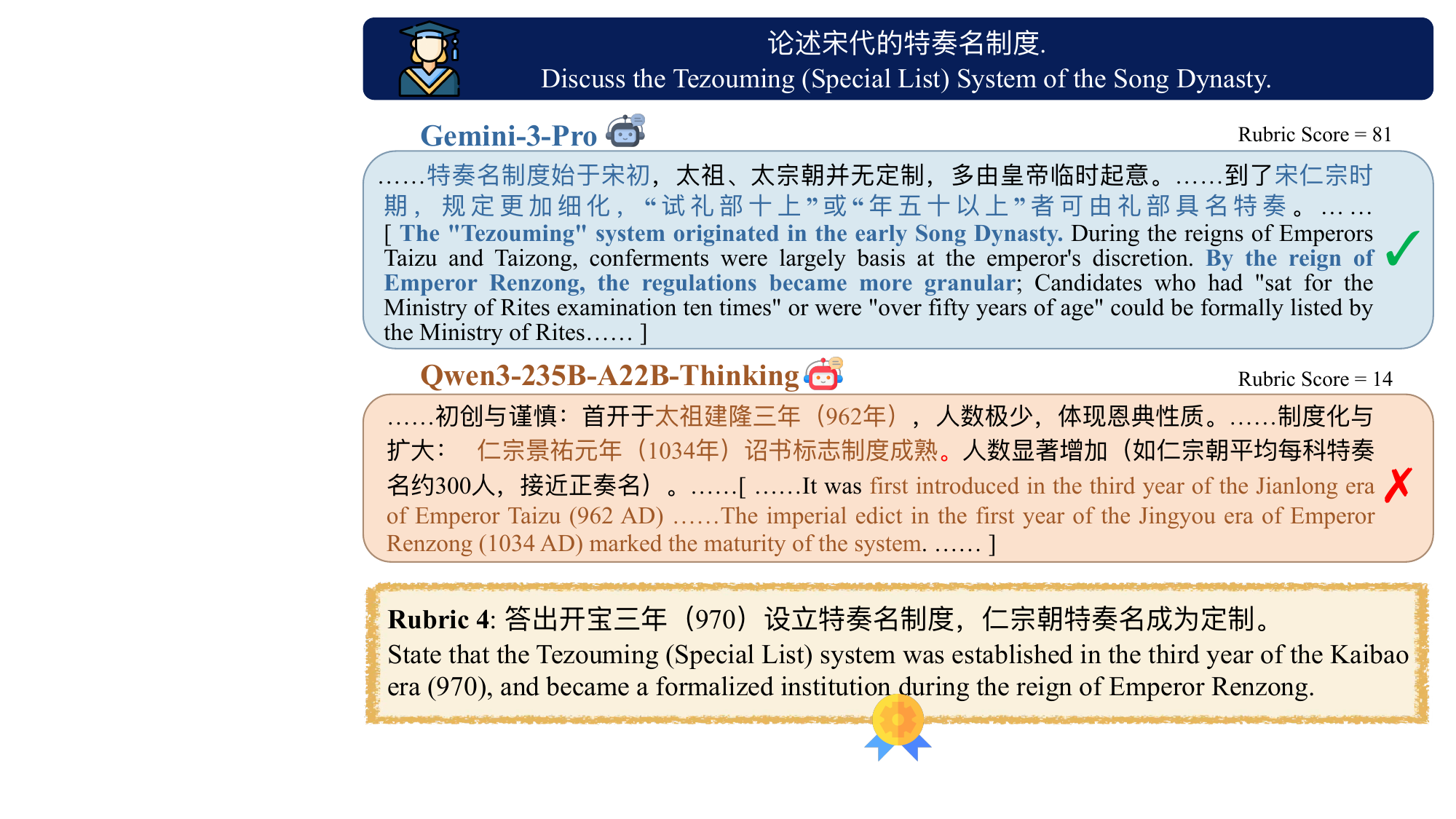}
    \caption{
    An illustration of LLMs’ factual hallucinations in historical research tasks. Severe hallucinations persist in specialized historical contexts, even among advanced LLMs.
    }
    \label{fig:intro}
\end{figure}

Recent breakthroughs in LLMs~\cite{openai2024gpt4technicalreport,yang2025qwen3} profoundly impact historical research by automating the processing of vast digitized archival collections~\cite{chenobibench} and facilitating narrative generation~\cite{ye2025llms4all}. Such paradigm shift motivates the development of historical benchmarks for LLMs. Existing evaluations cover various dimensions, including general historical knowledge~\cite{hauser2024largeHiST-LLM}, ancient languages~\cite{zhou2023wyweb,piryani2024chroniclingamericaqa}, and multi-modal historical materials~\cite{shi2023m5hisdoc,ghaboura2025time,zhangms-bench2025}, providing the evaluation of LLMs' comprehension of general history and their ability to process complex historical materials.

However, a significant gap remains between basic historical knowledge and professional historical research. As shown in Figure~\ref{fig:intro}, LLMs still suffer from hallucination problems and fail to resolve conflicting historical records, which are also demonstrated in recent studies~\cite{zhou-etal-2025-hanfu,ghaboura2025time}. Rather than merely retrieving facts, professional historical research hinges on complex skills such as socio-political contextualization, evidentiary reasoning and interpretive synthesis~\cite{Weinstein2005HistoryWA,shiyuanxue}. Yet current benchmarks largely overlook the evaluation of such higher-order skills, leaving LLM capabilities on genuinely professional historical tasks both underexplored and easy to overestimate.

To bridge this gap, we propose \textbf{\NAME}, the first \textbf{Pro}fessional \textbf{Hist}ory \textbf{Bench}mark framework.
Through deep interdisciplinary collaboration between AI researchers and professional historians, \NAME{} is anchored in the \textit{Chinese Imperial Examination} (\keju, \zh{科举}), a civil service selection system that operated for over 1,300 years and profoundly influenced the East Asian civilizations~\cite{Weinstein2005HistoryWA,Chang1942ChinaAE}. 
Centered on \keju{}, historians curated 400 expert-level, challenging questions and, crucially, handcrafted fine-grained scoring rubrics tailored to each question, amounting to 10,891 rubrics in total. These criteria systematically evaluate 9 core historical research capabilities (e.g., evidentiary reasoning, temporal reframing), ensuring comprehensive and rigorous professional assessment. Our systematic evaluation and analysis of 18 advanced LLMs reveals significant deficiencies in professional historical research capabilities, identifying specific weaknesses in handling complex, conflicting, and context-dependent historical tasks.

In summary, our contributions are as follows:
\begin{itemize}[topsep=2pt,itemsep=1pt,parsep=0pt,partopsep=0pt]
    \item  We construct 400 challenging history questions written by historians, and for the first time use the \textit{Keju} as a distinctive lens to cover roughly 1,300 years of historical evolution in ancient China, forming a multi-level benchmark that spans from basic factual understanding to historical research tasks.
    \item  We further propose the first history evaluation framework for LLMs: historians manually created 10,891 scoring rubrics covering nine categories of historical research abilities, and we also designed a complete rubric-based evaluation pipeline to enable systematic and reproducible assessment of model performance.
    \item We conducted a systematic evaluation of 18 advanced LLMs. Our results reveal significant deficiencies in current models' ability to perform professional historical research.
\end{itemize}

\section{Related Work}
\subsection{MLLM Evaluation for History}
In the multimodal domain, researchers have established benchmarks evaluating cross-cultural understanding~\cite{Liu2021VisuallyGRmarvlbench,chiu-etal-2025-culturalbench,Vayani2024AllLMbench}, Chinese visual culture~\cite{zhang-etal-2025-mllms-ciibench,zhou-etal-2025-hanfu}, and historical material processing~\cite{shi2023m5hisdoc,chenobibench,ghaboura2025time,liu-etal-2025-mcsbench,zhangms-bench2025}. However, these benchmarks primarily focus on basic perceptual tasks such as visual recognition and OCR, largely neglecting the deep reasoning and argumentation capabilities essential for professional historical research.

\subsection{LLM Evaluation for History}

A burgeoning body of evaluation research has explored the application of LLMs within historical research. Existing work covers global history knowledge (\textbf{HiST-LLM}~\cite{hauser2024largeHiST-LLM}, \textbf{HiBenchLLM}~\cite{chartier2025hibenchllm}), as well as historical document understanding (\textbf{M5HisDoc}~\cite{shi2023m5hisdoc}, \textbf{ChroniclingAmericaQA}~\cite{piryani2024chroniclingamericaqa}, \textbf{AC-EVAL}~\cite{wei2024aceval}, and \textbf{C$^{3}$Bench}~\cite{cao2024c}).  For \textbf{general Chinese capabilities}, benchmarks such as \textbf{C-Eval}~\cite{Huang2023CEvalAM}, \textbf{CMMLU}~\cite{Li2023CMMLUMM}, and \textbf{Chinese SimpleQA}~\cite{He2024ChineseSimpleQA} provide comprehensive assessments of fundamental knowledge and QA across diverse disciplines. In the specialized field of \textbf{Classical Chinese Studies (CCS)}, \textbf{ACLUE}~\cite{zhang-li-2023-large-ACLUE} and \textbf{C-CLUE}~\cite{Ji2021CCLUEAB} target tasks ranging from ancient text comprehension to Named Entity Recognition, while \textbf{WYWEB}~\cite{zhou2023wyweb} and \textbf{WenMind}~\cite{Cao2024WenMindAC} expand the scope to include sequence labeling, machine translation, and holistic alignment with human intuition.
While existing benchmarks effectively assess general knowledge and linguistic proficiency, they fail to capture specialized capabilities such as low-resource context adaptation and deep historical reasoning.

Compared with existing work, the uniqueness and advantages of \NAME{} are twofold. First, ProHist-Bench is the first benchmark focused on in-depth research into the history of \keju{}, moving beyond basic historical QA toward specialized historical research. Second, we establish a more rigorous and fine-grained evaluation framework for assessing research-oriented abilities, including fact organization, historical comparison, and evidentiary reasoning. This framework also provides a valuable reference for future benchmark construction in other historical domains.

\begin{table*}[t]
\centering
\small
\setlength{\tabcolsep}{1pt}
\renewcommand{\arraystretch}{1.1}
\begin{tabular}{m{0.10\textwidth} m{0.20\textwidth} m{0.65\textwidth}}
\toprule
\textbf{Task ID} & \textbf{Task Name} & \textbf{Task Description} \\
\midrule
T1 & Term Interpretation & \textbf{Definition:} Assesses the model's ability to understand and explain historical terms. \newline \textbf{Example:} Explain the term \textit{Gongshi} (graduate of the \keju{} examination). \\
\midrule
T2 & Fact QA & \textbf{Definition:} Assesses the model's ability to organize and present historical facts. \newline \textbf{Example:} Briefly explain how the Qing dynasty's entry inspection system prevented cheating in the \keju{} examination. \\
\midrule
T3 & Historical Reasoning & \textbf{Definition:} Assesses the model's ability to make comparisons, integrate viewpoints, and reason with evidence based on historical facts. \newline \textbf{Example:} Discuss the \textit{Liudeng Chuzhi Fa} (Six-Grade System of Promotion and Demotion). \\
\midrule
T4 & \celun{} & \textbf{Definition:} Assesses the model's cross-contextual reasoning and decision-making abilities within a specific historical context. \newline \textbf{Example:} Assume you are a Qing dynasty candidate in the \keju{} examination in the 46th year of Qianlong (1749). Please write an essay in the Eight-Legged Essay (baguwen) format, following the specific rules for writing. If any characters need to be avoided due to taboos, use pinyin. Limit your response to 700 words. Provide only the essay and do not include any other content. The topic is: \begin{CJK}{UTF8}{gbsn}{孟子曰：待文王而后兴者，凡民也}\end{CJK} (Mencius said: Those who wait for Wenwang to rise are all the people.). \\
\bottomrule
\end{tabular}
\caption{Task definitions in \NAME{}. The benchmark is organized into four task types: Term Interpretation (T1), Fact QA (T2), Historical Reasoning (T3), and \celun{} (T4). }
\label{tab:task_definitions}
\end{table*}

\section{Dataset Construction}
\subsection{Design Principles}
Historical research presents distinctive reasoning challenges that extend beyond basic knowledge comprehension to encompass accurate terminology interpretation, evidence-based deduction, and the construction of coherent narratives grounded in historical context. \NAME{} aims to bridge the gap between comprehensive LLM evaluation and the intrinsic challenges of historical research. Aligned with the characteristics of historical scholarship, we adhere to the following principles in constructing \NAME{}.
\paragraph{Historical Representativeness.} We emphasize that historians are typically not generalists, but rather experts in a particular historical field, focusing on in-depth research of specific historical periods or events. The \keju{} examinations, as an important mechanism for talent selection in ancient China, spans over 1,300 years of history, and its study represents an extremely important direction for contemporary Chinese historical research. Our dataset contain a substantial amount of historical knowledge about \keju{}, and only by mastering this knowledge can historical researchers conduct in-depth research on \keju{} and thereby provide new insights. Although \keju{} research differs from other historical research fields in terms of research content, their research methodologies often share similarities. Therefore, examining LLMs' capabilities in fact organization, evidence reasoning, and other aspects through this topic can authentically reflect the performance level of LLMs when conducting in-depth historical research.

\paragraph{Historical Authority.} Historical authority is not an inherent or immutable property of any single narrative, but rather a defensible interpretation grounded in available evidence and established scholarly norms. Accordingly, drawing on a substantial body of extant historical materials and historiographical scholarship—such as \textit{Qingdai Zhujuan Jicheng} (\begin{CJK}{UTF8}{gbsn}{清代朱卷集成}\end{CJK}) and the latest academic findings—we develop a rigorously grounded and well-justified evaluation framework to assess the capabilities of LLMs.

\paragraph{Multi-dimensional Capability Assessment.} A well-designed benchmark should capture the diverse capabilities required for historical research. To this end, \NAME{} is constructed along four dimensions. First, in terms of \textbf{task diversity}, as shown in Table~\ref{tab:task_definitions}, \NAME{} draws inspiration from historical examinations and covers four task types: Term Interpretation (T1), Fact QA (T2), Historical Reasoning (T3), and \textit{Celun} (Policy Essay) Generation (T4). Second, in terms of \textbf{difficulty diversity}, all questions are annotated as either \emph{General} or \emph{Hard}, depending on whether they primarily require basic factual understanding or more advanced abilities such as multi-step reasoning, multi-source integration, and interpretive judgment. This binary scheme reduces subjective ambiguity and improves annotation consistency. Third, in terms of \textbf{content diversity}, \NAME{} spans a broad range of historical domains, including politics, economics, social life, transportation, and intellectual-cultural history. Fourth, in terms of \textbf{temporal coverage}, it covers more than 1,300 years across nine dynasties, enabling evaluation across varied historical periods.

\subsection{Authoritative Data Sources}
\NAME{} is constructed based on 125 core references spanning three categories (see Appendix~\ref{app-reference} for the complete bibliography):
\textbf{(1) Ancient Chinese Text.} 
Qing dynasty official documents, local gazetteers, and examination archives, including \textit{Qingdai Zhujuan Jicheng} (\begin{CJK}{UTF8}{gbsn}{清代朱卷集成}\end{CJK}), and \textit{Qingdai Keju Kaoshi Shulu} (\begin{CJK}{UTF8}{gbsn}{清代科举考试述录}\end{CJK}) by the last \textit{Tanhua} (\begin{CJK}{UTF8}{gbsn}{探花}\end{CJK}) (third-place scholar).
\textbf{(2) Authoritative Scholarly Monographs.} 
Two Chinese national comprehensive histories, alongside seminal works by leading scholars including Ichisada Miyazaki, Benjamin Elman, and others, ensuring global scholarly perspectives.
\textbf{(3) Top-tier Academic Journals.} 
Cutting-edge research papers from CSSCI-indexed journals, including \textit{Historical Research}, \textit{Modern Chinese History Studies}, and \textit{Qing History Journal}, etc.

\begin{figure}[t]
    \centering
    \includegraphics[width=0.95\linewidth]{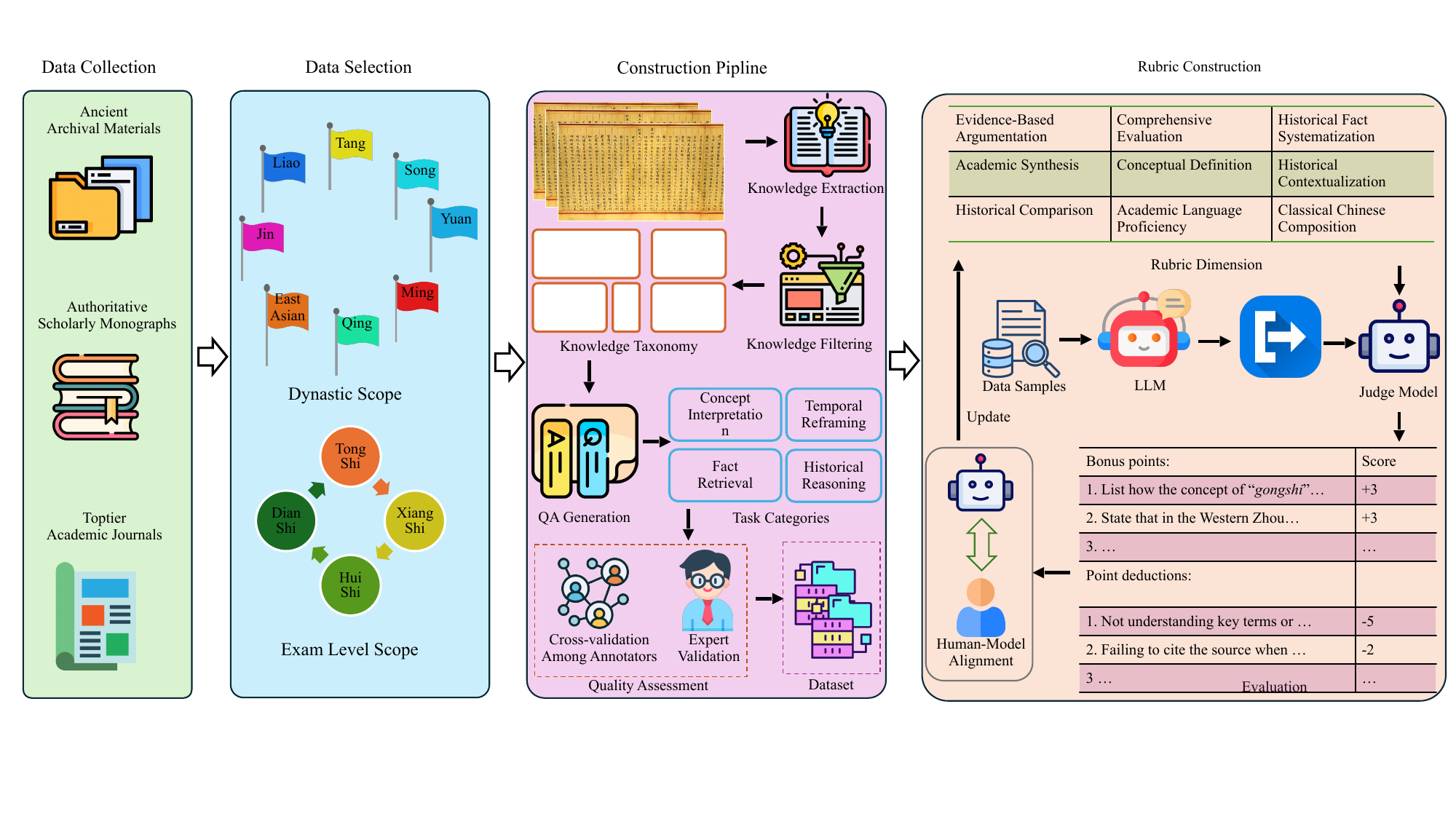}
    \caption{
   Overview of the dataset construction pipeline.
    }
    \label{fig:method}
\end{figure}

\begin{figure}[t]
    \centering
    \includegraphics[width=1.0\linewidth]{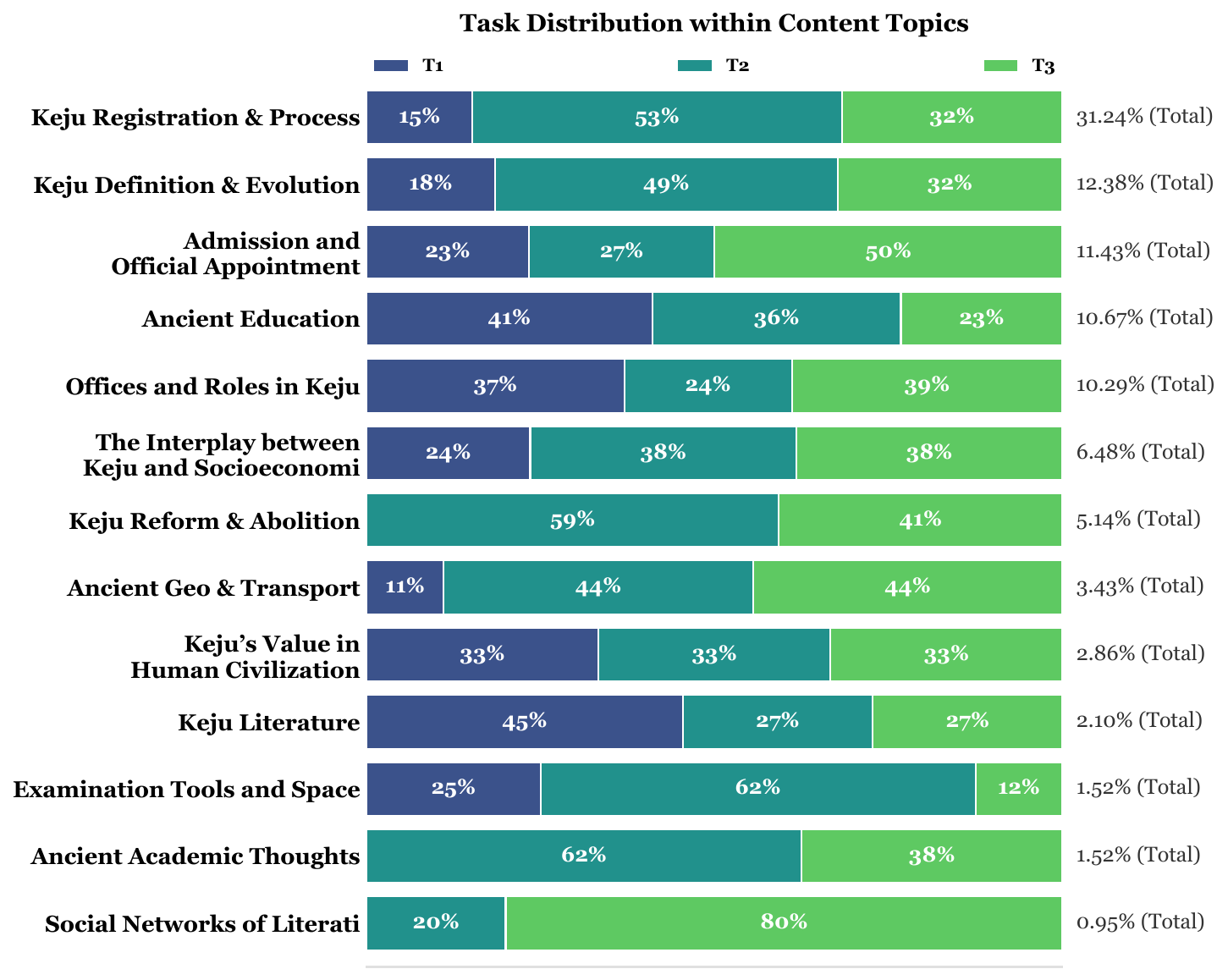}
    \caption{
    Distribution of task categories across various content topics. The chart displays the ratio of T1-T3 tasks for each topic.
    }
    \label{fig:top_statistics}
\end{figure}

\subsection{Expert Annotation}
The questions and reference answers in this dataset were all manually constructed by historians. The dataset construction pipeline is shown in Figure~\ref{fig:method}.

\paragraph{Question Writing} Based on thirteen preset dimensions and temporal distributions, we extracted  related knowledge from historical materials and research papers, converting them into four question types: Term Interpretation, Fact QA, Historical Reasoning, and \celun{}. We introduced LLMs to assist historians in refining the question phrasing: identifying ambiguities through multi-round Q\&A testing, adjusting qualifiers and background prompts accordingly, and attempting different questioning styles to evaluate the models' diverse capabilities.

\paragraph{Answer Annotation} We consulted a total of 125 references to compile the answers and provided detailed citations for each one. The answers underwent cross-verification via multi-source historical materials, adhering to mainstream consensus while listing various viewpoints for unsettled issues, emphasizing argumentative logic rather than trivial details. It is worth noting that, although we attempted to use LLMs to assist in answer annotation, we found that LLMs exhibit significant shortcomings in subject completion within ancient texts, terminology understanding, and the mechanisms of historical events. Therefore, ultimately, all reference answers were written sentence-by-sentence by historians to ensure historical accuracy. 

\paragraph{Quality Assessment} 
To ensure data quality, we adopt a four-stage quality-control pipeline. First, six historians independently wrote 100 questions each based on 125 literatures and annotated reference answers, rubrics, and metadata such as dynasty and topic. Second, each question was reviewed by at least one non-original historian, and any disagreements were adjudicated by a third senior historian, ensuring at least two-person review for every question. Third, two senior historians independently checked all questions, and any remaining disagreements were resolved by majority voting, providing full-coverage and high-standard quality control. Fourth, we randomly sampled 5\% of the data for in-depth review by historians. If any error was found, a full recheck was triggered, followed by repeated sampling until the sampled batch reached 100\% accuracy. In addition, we discarded questions that were too easy, overly subjective, or highly contested such that stable criteria could not be defined within the designated source boundary, thereby ensuring the executability and consistency of the evaluation standards.

\subsection{Dataset Statistics}

\NAME{} consists of 400 expert-curated historical research questions spanning four core tasks: Term Interpretation, Fact QA, Historical Reasoning, and \celun{}. As shown in Figure~\ref{fig:top_statistics}, the dataset covers 13 topics, with Registration and Process accounting for the largest share (31.24\%), followed by Definition and Evolution (12.38\%) and Entry into Officialdom and Appointments (11.43\%). In terms of task distribution, T2 constitutes the largest proportion (approximately 43\%), followed by T3 (approximately 35\%) and T1 (approximately 22\%). More fine-grained statistics by task type, difficulty, and dynasty are provided in Appendix~\ref{sec:app-topic}.

\begin{table}[t]
    \centering
    % \small
    \renewcommand{\arraystretch}{1.1}
    \setlength{\tabcolsep}{1mm}
    \resizebox{0.98\linewidth}{!}{% 
\begin{tabular}{cccc}
  \toprule
  ID &\textbf{Capability  Category}  & \textbf{Count} & \textbf{Percentage} \\
  \midrule
R1 & Concept Definition & 1,002& 9.20\%  \\
R2& Fact Organization  & 6,452& 59.24\% \\
R3& Historical Comparison & 173& 1.59\%  \\
R4& Evidentiary Reasoning & 732& 6.72\%  \\
R5& Comprehensive Evaluation& 315& 2.89\%  \\
R6& Viewpoint Integration & 97 & 0.89\%  \\
R7& Academic Expression& 1,167& 10.72\% \\
R8 & Classical Writing  & 91 & 0.84\%  \\
R9& Temporal Reframing & 862& 7.91\%  \\
  \midrule
  Total & /& 10,891& 100\%\\  
  \bottomrule
\end{tabular}
    }
    \captionsetup{justification=justified, singlelinecheck=false}
    \caption{
    Statistics of specific rubric dimensions (R1–R9) in \NAME{}.
    }
    \label{tab:rubric}
\end{table}

\section{Rubric Evaluation Framework}
To systematically evaluate whether LLMs can function like historians, we designed a multi-dimensional rubric evaluation framework, decomposing LLM performance into 9 observable dimensions, defining clear levels and rules for each dimension, thereby enhancing scoring consistency and reproducibility.
\subsection{Rubric Definition}
A rubric is a set of criteria or scoring rules developed by historians to evaluate the quality of responses to historical questions. As shown in Table~\ref{tab:rubric}, we design nine rubric dimensions to capture different aspects of historical research ability. For T1--T3, the rubric covers \textbf{Concept Definition} (R1), \textbf{Fact Organization} (R2), \textbf{Historical Comparison} (R3), \textbf{Evidentiary Reasoning} (R4), \textbf{Comprehensive Evaluation} (R5), \textbf{Viewpoint Integration} (R6), and \textbf{Academic Expression} (R7). Together, these dimensions assess whether a model can accurately define historical concepts, organize and compare facts, reason with evidence, synthesize scholarly perspectives, and present arguments in a rigorous academic manner. For T4 (\celun{}), we further introduce \textbf{Classical Writing} (R8) and \textbf{Temporal Reframing} (R9), which evaluate the model’s ability to generate historically grounded responses under dynasty-specific literary, institutional, and stylistic constraints. To preserve historical fidelity, the rubric also imposes explicit penalties for major violations of period conventions, such as departures from genre norms or the use of taboo expressions. Detailed definitions, scoring rules, and penalty criteria are provided in Appendix~\ref{sec:app-postive-rubric}.

Each dimension is weighted according to its importance for historical research and the level of capability it demands from the model. Basic dimensions, such as Concept Definition and Fact Organization, mainly assess knowledge recall and are therefore assigned lower weights. By contrast, more advanced dimensions, such as Evidentiary Reasoning and Temporal Reframing, require broader knowledge and more complex reasoning, and thus receive higher weights. The nine dimensions are applied selectively across tasks: T1--T3 emphasize R1--R7, while T4 focuses on the specialized abilities captured by R8 and R9.

\paragraph{Penalty Rubrics.} In addition to the positive evaluation rubrics designed for the nine capabilities above, we further establish a set of universal penalty rubrics applicable to all tasks, aimed at capturing intolerable errors, hallucinations, and other problematic behaviors in LLM-based historical research. Detailed definitions are provided in Appendix~\ref{app-negrubric}.

\subsection{Rubric Construction}
Constructing fine-grained and objective rubrics is challenging, but necessary for in-depth evaluation of LLMs in history. Existing benchmarks mainly rely on automated metrics such as Accuracy and BLEU, which capture surface-level textual similarity but fail to assess higher-order historical abilities, such as fact organization, comparative analysis, and evidence-based interpretation.

Rubric-based evaluation provides a more suitable alternative. While this approach has been validated in expert domains such as law~\cite{DBLP:journals/corr/abs-2601-16669} and healthcare~\cite{DBLP:journals/corr/abs-2505-08775}, our work is, to the best of our knowledge, the first to apply it to history. All rubrics in our benchmark are question-specific and developed by historians. For T1--T3, we adopt an \textbf{Iterative Refinement} strategy, where historians draft initial criteria and iteratively revise them based on real model outputs to avoid overly coarse judgments. For T4, we follow a \textbf{Historical Authenticity} principle, grounding the rubrics in the \textit{Qing Dynasty Imperial Examination Regulations} (\zh{清代科场条例}) and relevant scholarship to ensure historical fidelity and temporal accuracy. To enhance objectivity, we invited multiple historians to translate historiographical requirements into quantifiable criteria and conducted repeated cross-validation, reducing individual bias and strengthening disciplinary rigor.

\subsection{Rubric Statistics}
As shown in Tables~\ref{tab:rubric} and~\ref{tab:rubric2}, \NAME{} comprises 10,891 fine-grained evaluation criteria, with an average of 27.23 criteria per question. T1-T3 tasks contain 9,938 criteria (91.25\%), while T4 task include 953 criteria (8.75\%). Among the nine capability categories, Fact Organization accounts for the largest proportion (59.24\%), reflecting its fundamental role in historical research, followed by Academic Expression (10.72\%) and Concept Definition (9.20\%). The distribution demonstrates our rubric criteria emphasis on core historical competencies while maintaining comprehensive coverage of diverse historical research capabilities.

\begin{table}[t]
    \centering
    % \small
    \renewcommand{\arraystretch}{1.1}
    \setlength{\tabcolsep}{2mm}
    \resizebox{0.7\linewidth}{!}{% 
\begin{tabular}{ccc}
  \toprule
  \textbf{Task} &  \textbf{Total \# Criteria} & \textbf{Avg. \# Criteria} \\
  \midrule
  T1 &1,650&18.33  \\
  T2 &3,963&26.42 \\
  T3 &4,325&36.04 \\
  T4 &953&23.83\\
  \midrule
  Total &10,891&27.23\\  
  \bottomrule
\end{tabular}
    }
    \captionsetup{justification=justified, singlelinecheck=false}
    \caption{
    Statistics of fine-grained evaluation criteria across different task categories.
    }
    \label{tab:rubric2}
\end{table}

\begin{figure}[h!]
    \centering
    \includegraphics[width=0.85\linewidth]{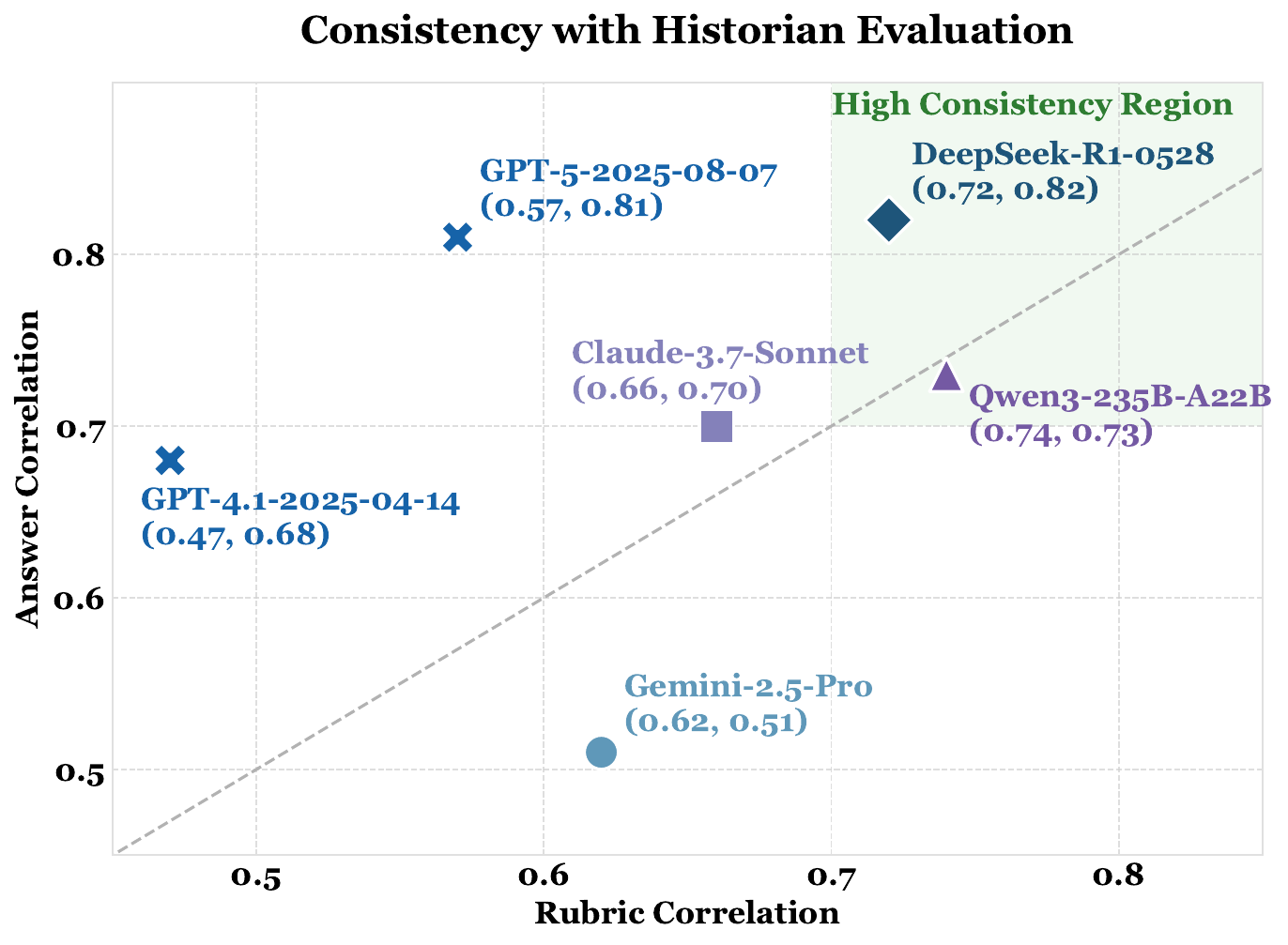}
    \caption{Consistency score of candidate judge models with manually crafted (human) ground truth (answers and rubric).
    % Human-consistency scores (\%) for different judge models under two evaluation settings (``Rubric'' and ``Answer''). 
    }
    \label{fig:consistency}
\end{figure}
\subsection{Evaluation Protocol}

We adopt an LLM-as-a-Judge paradigm to automate the evaluation process. To ensure reliable evaluation, we conducted a preliminary study comparing six candidate LLMs against expert historian evaluations. We randomly sampled 50 instances and collected the outputs of the evaluated LLMs. We asked each candidate judge model and human experts to perform item-by-item 0/1 hit annotation according to the rubric, obtaining fine-grained scores (rubric-level) and an aggregated total score (answer-level). We then computed the Pearson correlation coefficients between each candidate judge model and the experts on both types of scores. 
As illustrated in Figure~\ref{fig:consistency}, DeepSeek-R1 demonstrated the highest average consistency (0.77) with historians at both the answer level and the fine-grained rubric level, and was therefore selected as the judge model.

For scoring, we define a quantitative metric named \textbf{Rubric Score (RS)}. The judge model determines the presence (binary 0/1) of each criterion defined in the rubric. The final score is calculated by aggregating the weighted scores of all triggered items (including both bonus and penalty terms) and normalizing by the total potential positive score:
\begin{equation}
    \text{RS} = \max \left( 0, \frac{\sum (I_{b} \cdot w_{b}) + \sum (I_{p} \cdot w_{p})}{\sum w_{b}} \right)
\end{equation}
where $I_b$ and $I_p$ denote the indicators for bonus and penalty rubric items, respectively ($I \in \{0, 1\}$ denotes the indicator of whether a specific criterion is met), and $w$ represents the pre-defined score of that criterion. Negative scores are clipped to 0 to ensure a valid range.

\begin{table}[t]
    \centering
    \large
    \renewcommand{\arraystretch}{1}
    \resizebox{1.0\linewidth}{!}{% 
        \begin{tabular}{lcccc
        }
        \toprule
         \textbf{Model} &{\textbf{BL}} & {\textbf{RG}} & {\textbf{BS}} & {\textbf{RS}}\\
        \midrule
        \multicolumn{5}{c}{\cellcolor[HTML]{EFEFEF}\textit{Closed-Source Models}} \\
        \midrule
        Claude-Sonnet-4.5-Thinking&2.53 &4.76 &71.49 &12.99\\
        GPT-5.2&3.50 &3.46 &71.48 &11.07\\
        GPT-5.2-Thinking&4.45 &4.58 &71.55 &14.08\\
        GPT-o3-2025-04-16&7.88 &3.65 &72.60 &14.66\\
        Gemini-3-Pro-Preview&1.94 &6.27 &73.97 &26.71\\
        Gemini-3-Pro-Preview-Thinking&2.35 &5.16 &73.92 &26.73\\
        Qwen3-Max&4.77 &6.64 &75.01 &17.71\\
        \midrule
        \multicolumn{5}{c}{\cellcolor[HTML]{EFEFEF}\textit{Open-Source Models}} \\
        \midrule
        Llama-4-Scout-17B-16E&2.59 &3.09 &72.68 &2.72\\
        gpt-oss-120b&1.27 &1.78 &70.18 &10.75\\
        gpt-oss-20b&\textit{Fail}&\textit{Fail}&\textit{Fail}&\textit{Fail}\\
        Kimi-K2-Thinking&3.62 &6.43 &73.20 &22.79\\
        GLM-4.6-Thinking&2.09 &5.11 &72.30 &24.32\\
        Qwen3-14B-Thinking&1.89 &3.61 &72.93 &11.61\\
        Qwen3-32B-Thinking&2.06 &3.97 &72.99 &13.89\\
        Qwen3-235B-A22B-Thinking&1.08 &5.22 &72.50 &28.14\\
        DeepSeek-V3.2&4.65 &6.20 &73.52 &18.77\\
        DeepSeek-V3.2-Thinking&4.91 &6.16 &73.41 &18.72\\
        DeepSeek-R1-0528&1.93 &6.60 &73.15 &26.87\\
        \bottomrule
        \end{tabular}
    }
    \captionsetup{justification=justified, singlelinecheck=false}
    \caption{
    Performance comparison of various LLMs on tasks T1-T3. The reported metrics include average BLEU (BL), ROUGE (RG), BERTScore (BS), and Rubric Score (RS).
    }
    \label{tab:main}
\end{table}

\section{Experiments}
\subsection{Setup}
\paragraph{Evaluated Models} We conducted extensive experiments on \NAME, evaluating 7 closed-source LLMs (Claude-Sonnet-4.5-Thinking, GPT-series, Gemini-series and Qwen3-Max) and 11 open-source LLMs (Llama-4-Scout-17B-16E, gpt-oss-series, Kimi-K2-Thinking, GLM-4.6-Thinking, Qwen3-series and DeepSeek-series). We used official APIs or standard deployment setups for all LLMs. To ensure deterministic evaluation, we fixed all hyperparameters and eliminated randomness. 

\paragraph{Metric} 
We employ different evaluation metrics tailored to specific task types. We utilize two complementary evaluation approaches: (1) \textit{automatic metrics} including \textbf{BLEU (BL)}~\cite{papineni-etal-2002-bleu}, \textbf{ROUGE (RG)}~\cite{lin-2004-rouge}, and \textbf{BERTScore (BS)}~\cite{zhang2019bertscore} to measure the similarity between LLM outputs and reference answers; and (2) \textit{expert-based evaluation} using \textbf{Rubric Score (RS)} to assess the historical research capabilities of LLMs. For T4 task, we rely solely on Rubric Score (RS) for evaluation, as these tasks require nuanced assessment of research quality that cannot be captured by automatic metrics.

\begin{figure*}[t]
    \centering
    \includegraphics[width=0.98\linewidth]{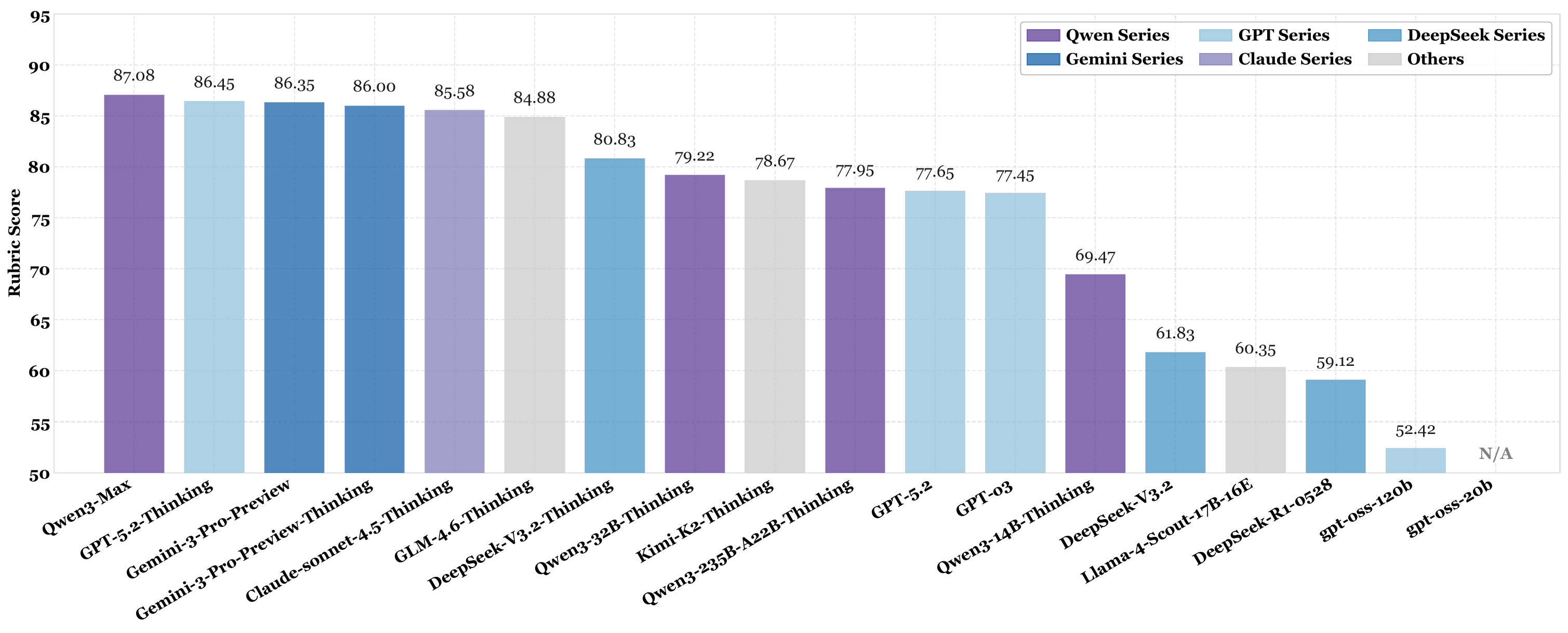}
    \caption{
    Main results of different models on the T4
    task. The table presents the performance of representative closed-source and open-source LLMs. The metrics
    reported are Rubric Score (RS).
    }
    \label{fig:t4res}
\end{figure*}

\subsection{Main results}

\paragraph{Core tasks in professional historical research remain extremely difficult for all current LLMs.}
As shown in Table~\ref{tab:main}, even top-performing LLMs such as Gemini-3-Pro and Qwen3-235B achieve RS scores barely approaching 30, while most LLMs score below 15. This pervasive underperformance indicates that existing LLMs still struggle to meet satisfactory professional historical research standards when handling tasks requiring high factual precision, complex logical reasoning, and nuanced understanding of specific historical contexts.

\paragraph{Chinese LLMs demonstrate superior performance in \NAME{}.}
Qwen3-Max achieves the strongest overall performance with an average BS of 75.01 and secures the top score of 87.08 on T4 task (Figure~\ref{fig:t4res}). Notably, open-source LLMs such as DeepSeek-R1-0528 (RS 26.87) follow closely behind, indicating that through large-scale parameters and strategic optimization, the open-source community has achieved capabilities approaching closed-source LLMs. Furthermore, LLMs trained on large-scale Chinese corpus, including GLM-4.6, Kimi-K2, and the DeepSeek series, consistently outperform LLMs like Llama-4-Scout (RS 2.72) and gpt-oss-120b (RS 10.75). This performance gap demonstrates the critical importance of domain-specific pretraining and cultural-linguistic alignment.

\begin{figure*}[h!]
    \centering
    \includegraphics[width=1.0\linewidth]{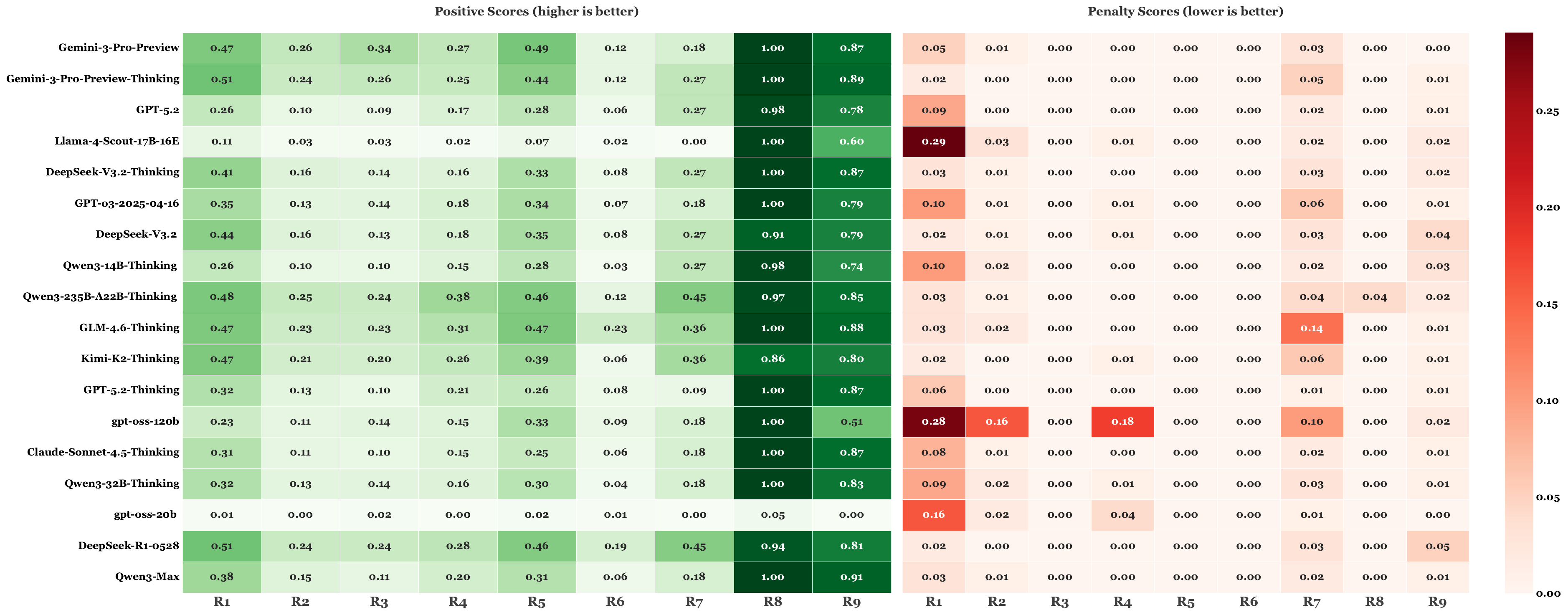}
    \caption{
    Performance heatmap across nine historical capability dimensions (R1-R9). \textbf{Left: Positive Rubric Hit Rate} (darker green indicates better performance). \textbf{Right: Penalty Rubric Hit Rate} (darker red indicates higher penalties), quantifying defects such as hallucinations and factual errors.
    }
    \label{fig:rubric_res}
\end{figure*}

\begin{table}[h!]
    \centering
    % \small
    % \renewcommand{\arraystretch}{1.1}
    \setlength{\tabcolsep}{1.2mm}
    \resizebox{0.9\linewidth}{!}{% 
        \begin{tabular}{l 
        *{4}{S[table-format=2.2, mode=text]}   % Level 1 的4列
        }
        \toprule
        \textbf{Model} & {\textbf{Role}} & {\textbf{Prof.}} & {\textbf{CoT}} & {\textbf{RAG}}\\
        \midrule
        \multicolumn{5}{c}{\cellcolor[HTML]{EFEFEF}\textit{Closed-Source Models}} \\
        \midrule
        Claude-Sonnet-4.5-Thinking& 19.42 &26.01&20.09&15.89\\
        GPT-5.2 &15.42&15.73&12.85&14.22\\
        GPT-5.2-Thinking &19.39&19.90&16.96&16.81\\
        GPT-o3 &17.09&17.59&14.33&20.29\\
        Gemini-3-Pro-Preview &32.94&30.95&25.08&26.53\\
        Gemini-3-Pro-Preview-Thinking &32.56&31.08&26.26&26.76\\
        Qwen3-Max &24.95&25.05&23.71&20.86\\
        % \cmidrule(r){2-17}
        Average&23.11&23.76&19.90&20.19\\
        \midrule
        \multicolumn{5}{c}{\cellcolor[HTML]{EFEFEF}\textit{Open-Source Models}}\\
        \midrule
        Llama-4-Scout-17B-16E-Instruct &5.11&4.55&3.88&5.88\\
        gpt-oss-120b &10.61&13.42&10.73&11.71\\
        gpt-oss-20b &0.41&0.77&0.19&2.79\\
        Kimi-K2-Thinking&31.61&36.10&34.10&31.18\\
        GLM-4.6-Thinking &31.73&32.51&27.78&24.10\\
        Qwen3-14B-Thinking &13.12&12.66&11.23&11.92\\
        Qwen3-32B-Thinking &17.23&15.19&15.25&13.33\\
        Qwen3-235B-A22B-Thinking &32.79&34.67&29.72&27.92\\
        DeepSeek-V3.2 &28.76&27.72&24.01&27.86\\
        DeepSeek-V3.2-Thinking &29.13&26.68&25.66&26.46\\
        DeepSeek-R1-0528&30.92&26.64&25.96&30.97\\
        % \cmidrule(r){2-17}
        Average&21.04&20.99&18.96&19.46\\
        \bottomrule
        \end{tabular}
    }
    \captionsetup{justification=justified, singlelinecheck=false}
    \caption{
    Impact of different prompting strategies on LLM performance. We evaluated four strategies: Historian Role-playing (Role), Professional Prompting (Prof.), Chain-of-Thought (CoT), and Retrieval-Augmented Generation (RAG).
    }
    \label{tab:prompt}
\end{table}

\subsection{In-Depth Analysis}
We evaluated model performance across five key dimensions: prompting strategies, capabilities, difficulty, dynasties, and topics. Here, we highlight the two most critical findings. Detailed analyses of remaining dimensions are provided in Appendix~\ref{app-moreanalysis}.

\subsubsection{The Impact of Prompting Strategy}
To investigate the impact of distinct prompting strategies, we conducted a systematic evaluation on the ProHist-Bench. As shown in Table~\ref{tab:prompt}, results demonstrate that Professional Prompting and Role-Playing consistently outperform Chain-of-Thought (CoT). This suggests that for history-literate LLMs, activating expert identity is more effective than enforcing step-by-step reasoning. Conversely, RAG methods, typically effective in knowledge-intensive tasks, underperform in this evaluation due to the severe scarcity of high-quality ancient historical data in existing retrieval corpora. Retrieved document fragments often contain noise or insufficient relevance, disrupting rather than supplementing the model's reasoning. Detailed RAG experiments and analysis are provided in Appendix~\ref{app:rag}.
\begin{figure}[t]
    \centering
    \includegraphics[width=1.0\linewidth]{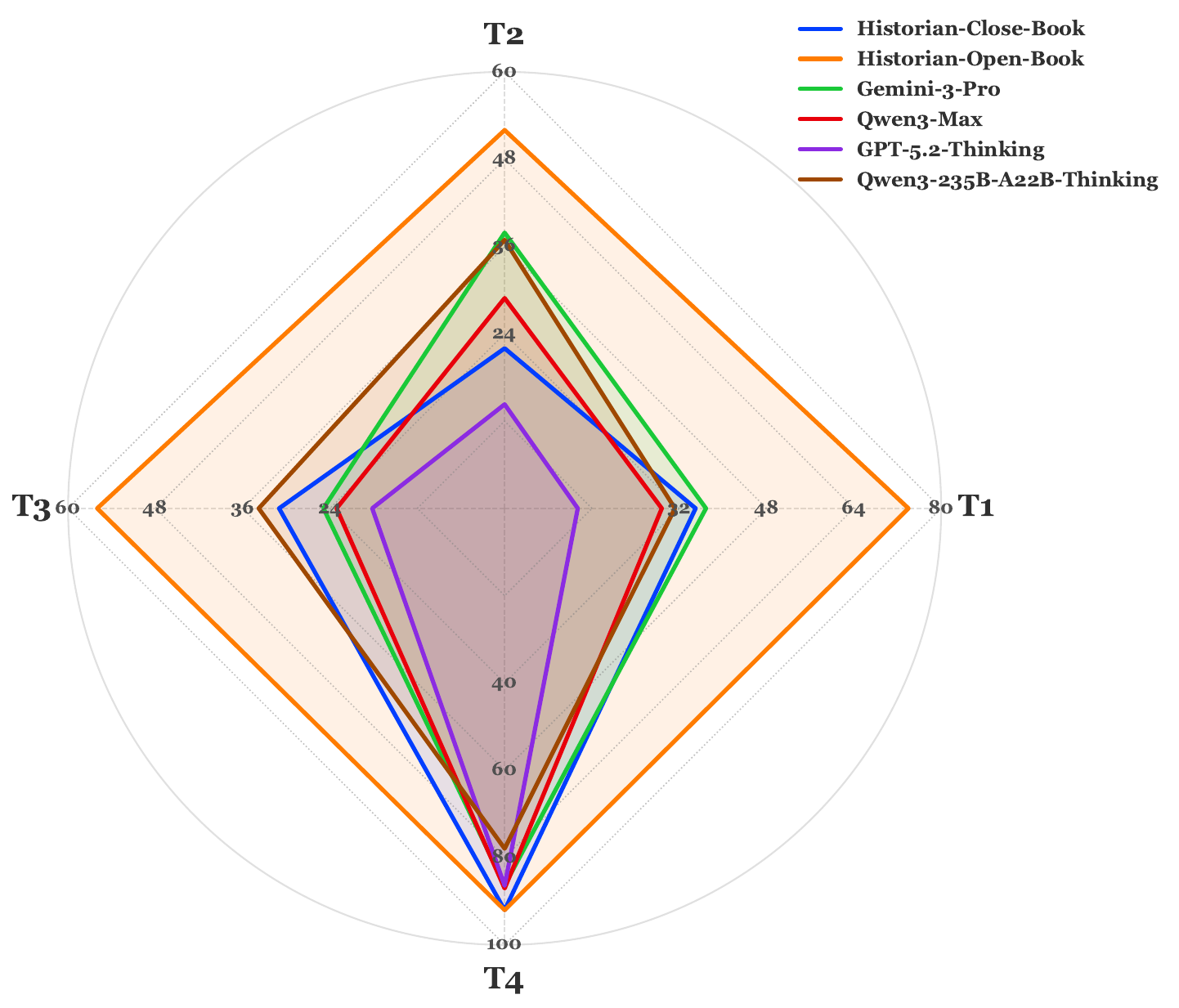}
    \caption{
    Performance comparison between human experts and LLMs.
    The Historian-Open-Book represents the expert ceiling with external
    resource access, while Historian-Close-Book serves as the unassisted human.
    }
    \label{fig:humanvsllm}
\end{figure}

\subsubsection{Fine-grained Capability Analysis}
To comprehensively evaluate the historical research capabilities of LLMs, we analyzed the distribution of Positive and Penalty Rubric Hit Rate (Figure~\ref{fig:rubric_res}), yielding several key observations: \textbf{i) Tool-like utility and formalization capabilities are mature}. High positive rubric hit rate in R8 (Classical Writing) demonstrate a mastery of rule-based logic. Furthermore, stable performance in R5 (Comprehensive Evaluation) confirms the models' ability to generate self-consistent narratives. 
\textbf{ii) Viewpoint Integration (R6) is currently the most prominent shortcoming of existing LLMs}. The R6 hit rate is extremely low across nearly all models (most are distributed between $0.02\sim0.12$, with the maximum only at $\sim 0.23$), indicating that regardless of model capability, there are significant deficiencies in their ability to resolve conflicting historical records. In addition, \textbf{iii) capabilities in historical comparison (R3), evidentiary  reasoning (R4), and fact organization are also generally weak}. Based on a detailed case study (see Appendix~\ref{app-case}), we find that the main reasons are twofold: first, most models are unable to distinguish the semantic evolution of the same concept across different historical periods; second, at the levels of reasoning and organization, although models can produce conclusions, they often lack a complete chain of historical evidence and support from a correct timeline, resulting in weak arguments.

\subsection{LLMs vs. Human}

To further assess LLMs in historical research, we compare SOTA LLMs with two human baselines: \textit{close-book historians}, representing internal knowledge only, and \textit{open-book historians}, representing a reference-augmented upper bound. The human baselines comprise around 20 history professors and PhD students. As shown in Figure~\ref{fig:humanvsllm}, SOTA LLMs achieve competitive performance on Historical Reasoning and \celun{}, but still lag far behind open-book historians on precision-critical tasks such as Term Interpretation and Fact QA. This gap suggests that, although LLMs have moved beyond basic historical QA, they have not yet reached the standard required for independent professional verification, and are better viewed as assistive tools rather than autonomous scholars.

\section{Conclusion}
In this paper, we introduced \NAME{}, an expert-curated benchmark anchored in \textit{\textit{Chinese Imperial Examination}} to rigorously evaluate the historical reasoning capabilities of LLMs. By transcending simple general knowledge QA to focus on professional historical research tasks, our interdisciplinary study reveals a significant proficiency gap between SOTA LLMs and professional historians. Our extensive evaluation highlights that current LLMs still struggle with the complex reasoning required for expert-level analysis.

\section*{Limitations}
We used a standardized prompt across all 18 models to ensure fairness and comparability. Although some models might perform better with model-specific prompt engineering, our goal was to assess their intrinsic historical reasoning under consistent conditions. Our results provide a robust baseline for performance in the historical domain.

\section*{Acknowledgement}
This work was supported by Ant Group.

\section*{Ethical Considerations}
Users of \NAME{} should recognize that both historical sources and annotations are inevitably shaped by the limitations of surviving materials and by historiographical interpretation. Accordingly, the dataset should be used with appropriate scholarly caution. \NAME{} is intended exclusively for non-commercial academic research, particularly for evaluating LLMs on historically grounded understanding and reasoning. All use must comply with applicable copyright regulations and research ethics standards, and any commercial or harmful use is strictly prohibited.

\bibliography{custom}

\appendix

\section{Positive Rubric Criteria}~\label{sec:app-postive-rubric}
To support fine-grained evaluation, we design nine rubric dimensions covering different aspects of historical understanding, reasoning, and generation. Detailed definitions are provided below.

\paragraph{R1. Concept Definition.} Assesses whether the model can provide concise conceptual definitions of historical phenomena before detailed elaboration, ensuring the definition captures core elements while maintaining historical accuracy. (2 points)
\paragraph{R2. Fact Organization.} As the most fundamental capability in historical research, this evaluates whether the model can clearly and comprehensively narrate the basic processes of events, institutions, phenomena, or intellectual developments from emergence through evolution to dissolution. (3 points)
\paragraph{R3. Historical Comparison.}  Building upon fact organization, this assesses the ability to identify historical evolution and cultural variations across different dynasties anchored in the \keju, demonstrating cross-temporal and cross-spatial comparative capabilities. (3 points)
\paragraph{R4. Evidentiary Reasoning.} Evidentiary Reasoning assesses the fundamental principle of arguments grounded in historical evidence by evaluating whether the model can derive reliable and academically consensual conclusions after organizing facts. This dimension ensures that all arguments are substantiated by credible historical sources rather than hallucinations and requires the model to cite concrete historical examples to support its viewpoints. (4 points)
\paragraph{R5. Comprehensive Evaluation.} Examines whether the model can provide holistic evaluations and interpretations of the significance of historical events, figures, and institutions—a fundamental capability. (1 point)
\paragraph{R6. Viewpoint Integration.} Evaluates whether the model can incorporate cutting-edge scholarship and research trends, present mainstream academic perspectives on historical phenomena, and establish clear connections among different scholars' viewpoints. (5 points)
\paragraph{R7. Academic Expression.} Assesses the rigor and professionalism of the model's academic writing. (5 points)
\paragraph{R8. Classical Writing.} Evaluates structural completeness based on essential elements of the eight-legged essay format, including opening (\zh{破题}), elaboration (\zh{承题}), transition (\zh{起讲}), and thesis (\zh{入题}).  (3 points)
\paragraph{R9. Temporal Reframing.} Evaluates the ability of LLM to reason within a specific historical context. Specifically, the rubric allocates 4 points to whether the writing style accords with the preferences of a given dynasty and 5 points to whether the essay meets basic compositional requirements. In the \celun{} task, the LLM is positioned as a candidate in the imperial examination of a particular dynasty, and the evaluation therefore also considers its grasp of the institutional context, literary conventions, and historical taboos (\zh{避讳词}) of the period. To preserve historical fidelity, we impose substantial penalties. Responses that depart from period style incur a 40 point deduction, whereas responses that violate examination norms incur a 60 point deduction. For example, the use of taboo words such as an emperor’s name could historically result in disqualification, and the 60 point penalty is intended to reflect the seriousness of such errors.

\section{General Penalty Rubric Criteria}~\label{app-negrubric}
General penalty rubric criteria includes:
\begin{itemize}
    \item[(1)] \textbf{Inappropriate Academic Formulation:} Instances of mechanically applying social science theories or making inappropriate analogies. \textit{\{-3 points\}}
    
    \item[(2)] \textbf{Fabrication of References or Materials:} Acts of fabricating cited references or inventing historical sources/materials. \textit{\{-5 points\}}
    
    \item[(3)] \textbf{Missing Citations:} Failure to provide standard citations when quoting Classical Chinese texts. \textit{\{-3 points\}}
    
    \item[(4)] \textbf{Chronological Conversion Errors:} Errors in converting ancient Chinese era names to Common Era (AD) years. \textit{\{-3 points\}}
    
    \item[(5)] \textbf{Core Concept Errors:} A lack of understanding or complete misinterpretation of specific historical terms. \textit{\{-5 points\}}
    
    \item[(6)] \textbf{Non-Academic Language Style:} The content is overly colloquial and lacks the rigor expected of academic writing. \textit{\{-1 points\}}
\end{itemize}

The criteria listed above apply to the entire evaluation process and are organized along two dimensions: i) Factuality Dimension (Items 1--5):
Designed to assess whether the model can accurately reflect historical records at the level of historical fact.
Items 1, 2, and 3 focus on detecting \textit{hallucinations}—instances where the model generates content that appears plausible and self-consistent but lacks a factual basis.
Items 4 and 5 focus on detecting the model's knowledge accuracy, checking for factual errors during the generation of specific knowledge points.
ii) Expression Dimension (Item 6):
Designed to assess the model's command of academic language. It requires that the generated content adheres to the academic norms of the humanities and social sciences, avoiding overly colloquial, entertaining, or casual unprofessional expressions.

\section{Dataset Scope and Topic Taxonomy}\label{sec:app-topic}
\begin{figure}[h!]
    \centering
    \includegraphics[width=1.0\linewidth]{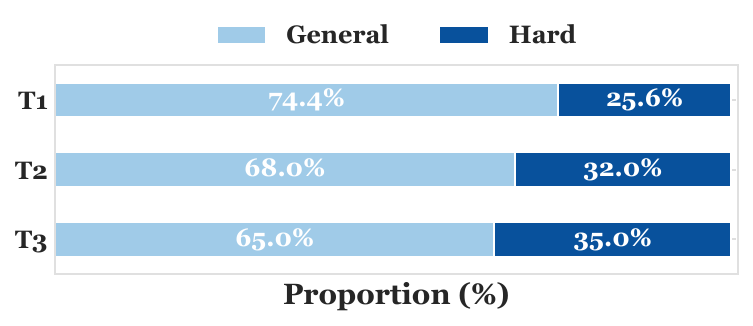}
    \caption{
   Distribution of difficulty levels (General vs. Hard) across different task categories.
    }
    \label{fig:diff_statistics}
\end{figure}

\begin{figure}[h!]
    \centering
    \includegraphics[width=1.0\linewidth]{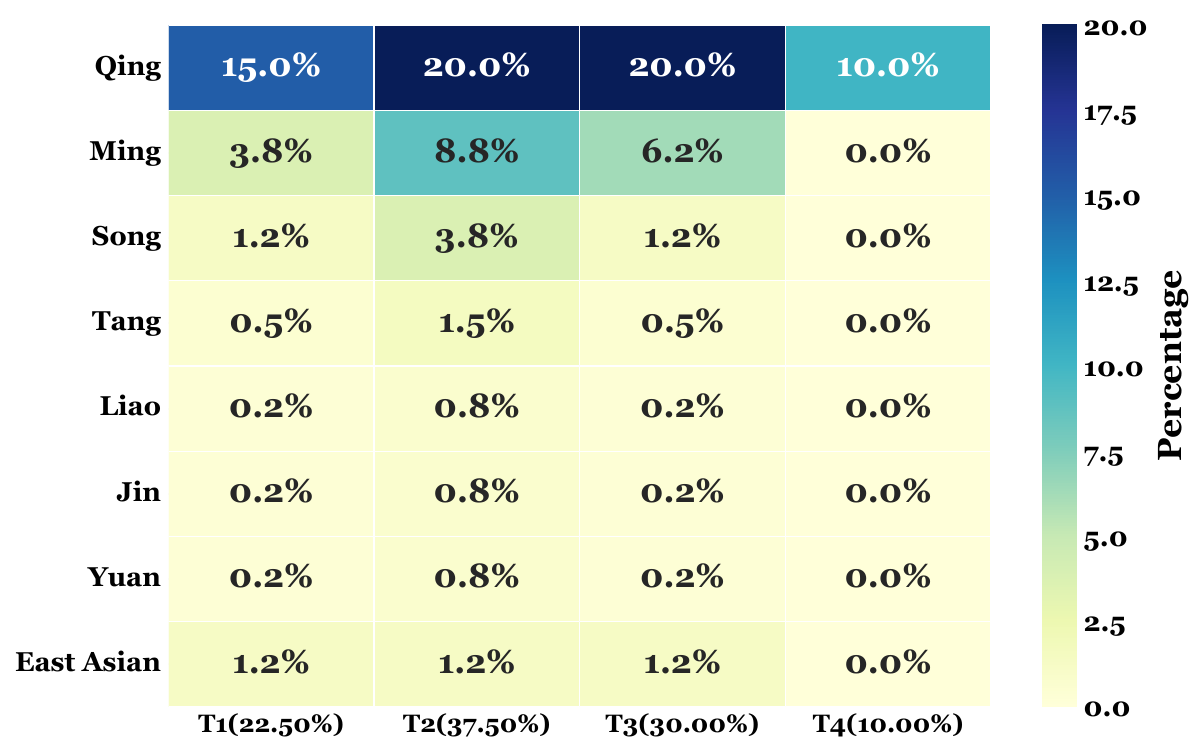}
    \caption{
    Cross-distribution of dynasties and task categories. Darker colors indicate higher proportions. ``East Asia'' represents historical knowledge of East Asian countries other than China.
    }
    \label{fig:dynasty_statistics}
\end{figure}

We present the statistical data of \NAME{} in Figures~\ref{fig:diff_statistics},~\ref{fig:dynasty_statistics} and~\ref{fig:top_statistics}.
\NAME{} consists of 400 expert-curated challenging historical research questions across four core tasks: Term Interpretation, Fact QA, Historical Reasoning, and \celun{}.
\subsection{Difficulty Distribution}
Difficulty distribution (Figure~\ref{fig:diff_statistics}): Historical Reasoning is the most challenging, with 35.0\% hard samples, compared to only 25.6\% for Term Interpretation.
\subsection{Task and Dynasty Distribution}
Task distribution (Figure~\ref{fig:dynasty_statistics}): Fact QA dominates at 37.50\%, while \celun{} is minimal at 10.00\%.
Historical period distribution (Figure~\ref{fig:dynasty_statistics}): The Qing Dynasty predominates, particularly in Term Interpretation and Historical Reasoning (both 20.0\%), due to its proximity to modern times and well-preserved documentation.
\subsection{Topic Taxonomy and Distribution}
\begin{figure*}[h!]
    \centering
    \includegraphics[width=1.0\linewidth]{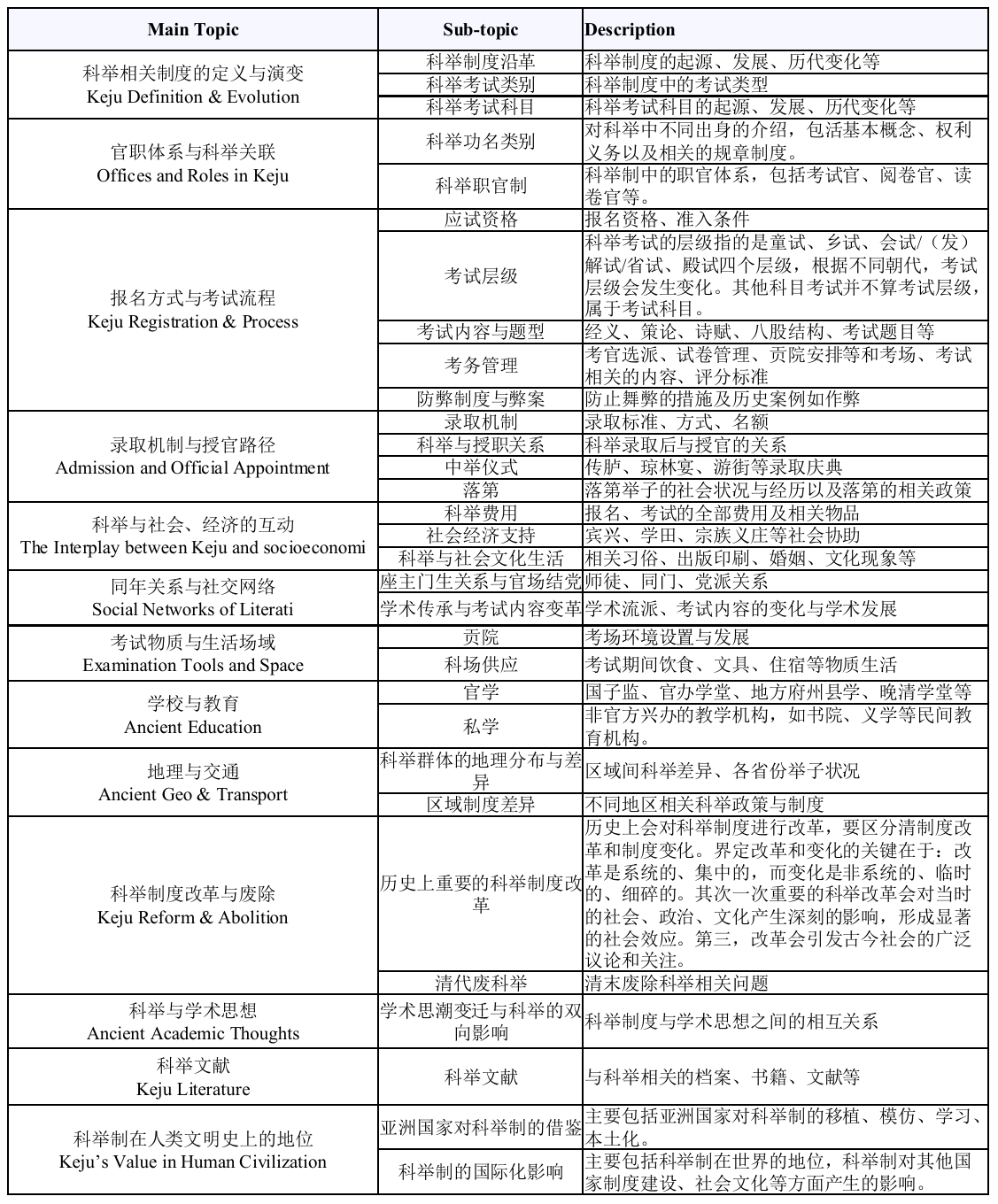}
    \caption{
    The hierarchical taxonomy of the topic framework for \NAME{}. This framework categorizes the dataset into 13 Main Topics and their corresponding Sub-topics, covering diverse dimensions ranging from institutional procedures and official roles to socio-economic interactions and global influences.
    }
    \label{fig:topic-tax}
\end{figure*}

To ensure a systematic and comprehensive evaluation of LLMs, we constructed a fine-grained knowledge taxonomy, as illustrated in Figures ~\ref{fig:topic-tax}. This hierarchical framework decomposes the complex historical domain into \textbf{13 Main Topics} and their corresponding \textbf{31 Sub-topics}. 

Unlike traditional benchmarks that focus primarily on factual recall, our taxonomy spans a broad spectrum of historiographical dimensions. It ranges from fundamental institutional procedures (e.g., \textit{Keju Registration \& Process}, \textit{Offices and Roles}) to complex socio-political interactions (e.g., \textit{The Interplay between Keju and Socio-economy}, \textit{Social Networks of Literati}). Furthermore, it incorporates high-level historical analysis, such as \textit{Keju Reform \& Abolition} and \textit{Keju's Value in Human Civilization}. This structured approach serves as the foundational guide for data collection, ensuring that the benchmark not only tests the model's knowledge retention but also its capability to handle diverse historical contexts and conduct deep reasoning across different granularity levels.

\section{Extended Discussion on Prompting Strategies}~\label{app-promptdetail}
\begin{figure*}[h!]
    \centering
    \includegraphics[width=1.0\linewidth]{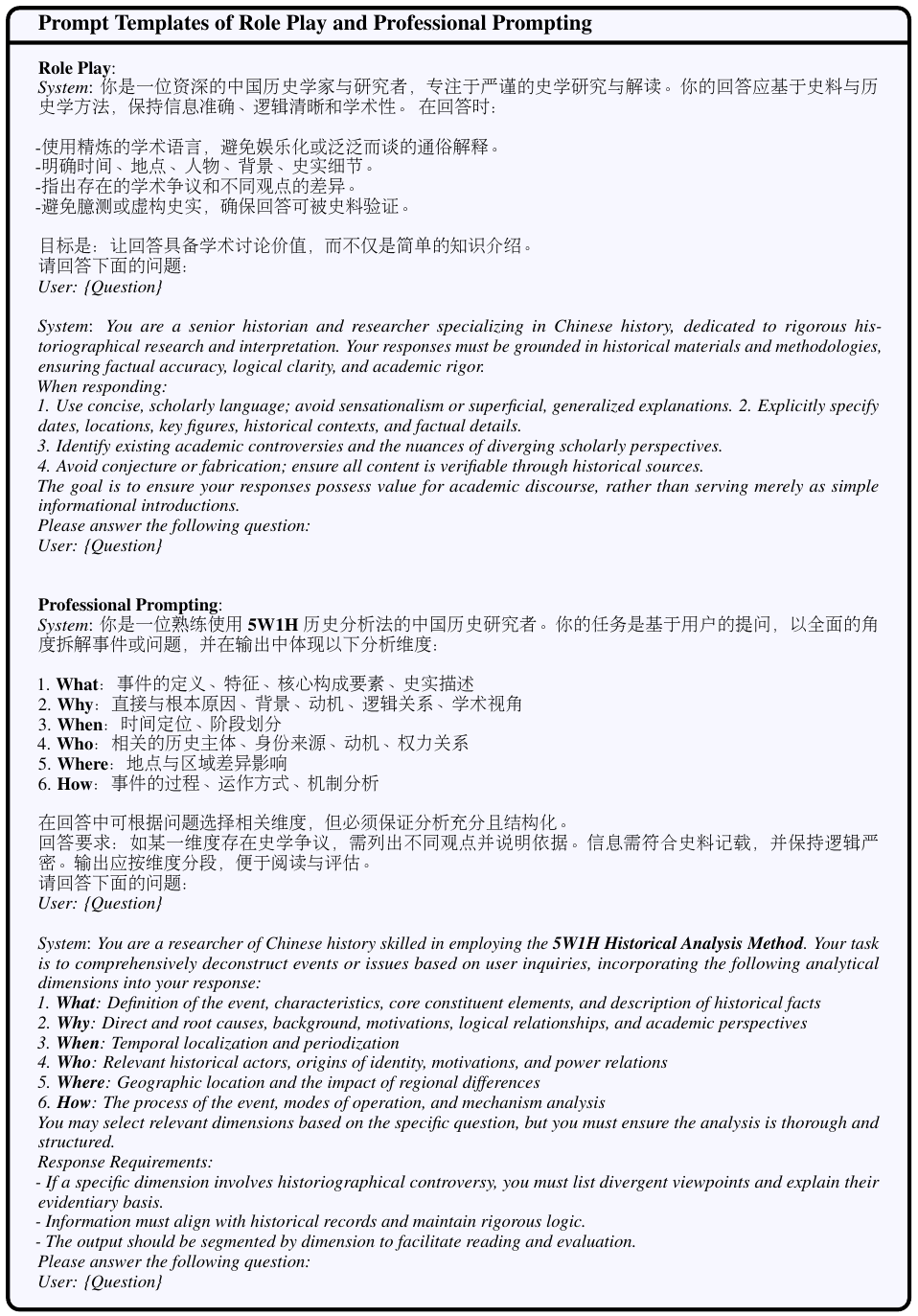}
    \caption{Prompt Templates of Role Play and Professional Prompting.
    }
    \label{fig:prompt-role}
\end{figure*}

\begin{figure*}[h!]
    \centering
    \includegraphics[width=1.0\linewidth]{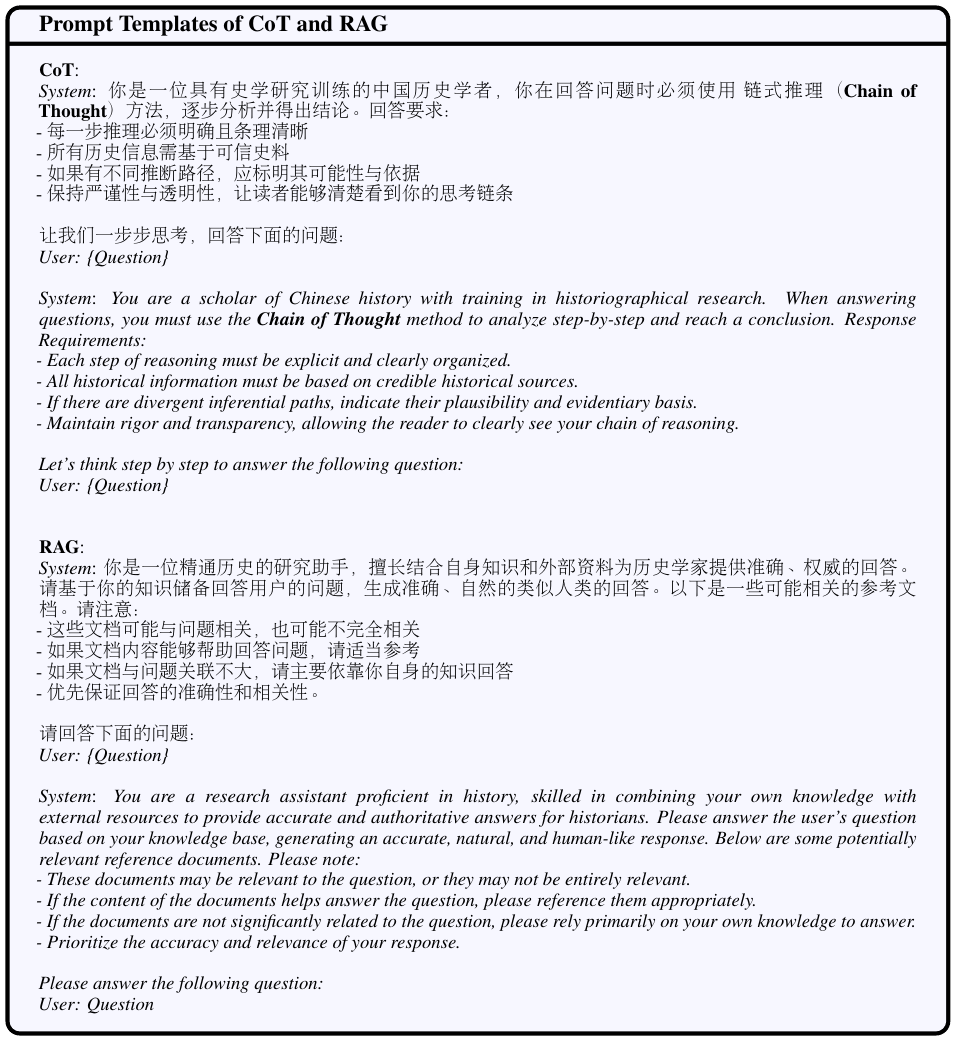}
    \caption{Prompt Templates of Role Play and Professional Prompting.
    }
    \label{fig:prompt-cot}
\end{figure*}

As illustrated in Table~\ref{tab:promptdetail}, we analyzed how different prompting methods—including Role, Prof., CoT, and RAG—affect the capabilities of various LLMs. Table~\ref{tab:promptdetail} details performance metrics across Term Interpretation, Fact QA, and Historical Reasoning, highlighting significant variations between closed-source and open-source LLMs under different prompting conditions. The specific prompt templates utilized in these experiments are provided in Figure~\ref{fig:prompt-role} and Figure~\ref{fig:prompt-cot}. 

\begin{table*}[t]
    \centering
    % \small
    \renewcommand{\arraystretch}{1.1}
    \setlength{\tabcolsep}{1.2mm}
    \resizebox{1.0\linewidth}{!}{% 
        \begin{tabular}{l 
        *{4}{S[table-format=2.2, mode=text]}   % Level 1 的4列
      !{\hspace{2mm}}                  % <--- 【关键】Level 1 和 2 之间的大间距
      *{4}{S[table-format=2.2, mode=text]}   % Level 2 的4列
      !{\hspace{2mm}}                  % <--- 【关键】Level 2 和 3 之间的大间距
      *{4}{S[table-format=2.2, mode=text]}   % Level 3 的4列
      !{\hspace{2mm}}                  % <--- 【关键】Level 3 和 Average 之间的大间距
      *{4}{S[table-format=2.2, mode=text]}   % Average 的4列
        }
        \toprule
        \multirow{2}{*}{\textbf{Model}} & 
      \multicolumn{4}{c}{\textbf{Term Interpretation}} & 
      \multicolumn{4}{c}{\textbf{Fact QA}} & 
      \multicolumn{4}{c}{\textbf{Historical Reasoning}} & 
      \multicolumn{4}{c}{\textbf{Average}} \\
      \cmidrule(r){2-5} \cmidrule(r){6-9} \cmidrule(r){10-13} \cmidrule{14-17}
        & {Role} & {Prof.} & {CoT} & {RAG} 
      & {Role} & {Prof.} & {CoT} & {RAG} 
      & {Role} & {Prof.} & {CoT} & {RAG} 
      & {Role} & {Prof.} & {CoT} & {RAG} \\
        \midrule
        \multicolumn{17}{c}{\cellcolor[HTML]{EFEFEF}\textit{Closed-Source Models}} \\
        \midrule
        Claude-Sonnet-4.5-Thinking &18.65&26.53&20.50&15.77&17.37&27.34&21.23&20.28&23.43&23.95&18.37&13.18&19.42&26.01&20.09&15.89\\
        GPT-5.2 &16.66&15.01&11.78&14.15&13.76&15.95&12.90&22.11&15.58&15.99&13.58&9.21&15.42&15.73&12.85&14.22\\
        GPT-5.2-Thinking &22.97&19.01&16.31&17.04&15.73&21.23&17.61&21.89&18.29&18.90&16.65&13.32&19.39&19.90&16.96&16.81\\
        GPT-o3 &18.03&19.23&16.13&17.12&13.27&18.84&15.39&30.11&20.62&14.80&11.67&16.93&17.09&17.59&14.33&20.29\\
        Gemini-3-Pro-Preview &31.95&37.64&31.20&27.88&29.19&30.83&24.51&28.44&39.59&26.08&21.21&24.04&32.94&30.95&25.08&26.53\\
        Gemini-3-Pro-Preview-Thinking &32.93&34.78&32.88&24.85&27.74&30.61&25.59&27.06&38.37&28.89&22.13&28.36&32.56&31.08&26.26&26.76\\
        Qwen3-Max &24.70&29.27&26.90&22.46&20.80&24.39&24.86&21.78&30.89&22.73&19.88&18.79&24.95&25.05&23.71&20.86\\
        \cmidrule(r){2-17}
        Average&23.70&25.93&22.24&19.90&19.69&24.17&20.30&24.52&26.68&21.62&17.64&17.69&23.11&23.76&19.90&20.19\\
        \midrule
        \multicolumn{17}{c}{\cellcolor[HTML]{EFEFEF}\textit{Open-Source Models}}\\
        \midrule
        Llama-4-Scout-17B-16E-Instruct &4.74&6.87&5.40&5.46&3.06&4.33&3.69&8.72&8.44&3.09&2.98&4.43&5.11&4.55&3.88&5.88\\
        gpt-oss-120b &12.75&10.58&8.12&10.08&8.56&16.79&13.20&16.89&9.80&11.35&9.61&9.89&10.61&13.42&10.73&11.71\\
        gpt-oss-20b &0.12&1.00&0.24&1.65&0.19&0.91&0.07&6.17&1.20&0.42&0.31&1.68&0.41&0.77&0.19&2.79\\
        Kimi-K2-Thinking &30.99&42.68&40.21&27.42&27.96&35.37&32.86&36.94&37.50&32.07&31.07&30.96&31.61&36.10&34.10&31.18\\
        GLM-4.6-Thinking &32.67&35.73&30.04&20.85&28.10&33.95&27.99&24.78&35.01&28.28&25.83&26.68&31.73&32.51&27.78&24.10\\
        Qwen3-14B-Thinking &13.01&15.89&12.62&11.62&11.07&11.31&10.49&17.06&16.03&11.93&11.09&8.89&13.12&12.66&11.23&11.92\\
        Qwen3-32B-Thinking &17.63&18.09&17.14&14.62&13.68&15.41&16.13&16.11&21.30&12.74&12.73&10.36&17.23&15.19&15.25&13.33\\
        Qwen3-235B-A22B-Thinking &33.29&38.93&34.79&28.92&29.23&36.12&30.75&31.11&36.69&29.67&24.63&24.93&32.79&34.67&29.72&27.92\\
        DeepSeek-V3.2 &28.25&31.32&28.69&26.31&24.22&28.37&25.29&29.06&35.69&24.20&18.89&28.54&28.76&27.72&24.01&27.86\\
        DeepSeek-V3.2-Thinking &29.41&33.66&31.53&23.69&24.02&25.25&25.46&30.78&35.46&23.23&21.50&26.25&29.13&26.68&25.66&26.46\\
        DeepSeek-R1-0528 &29.39&27.50&27.89&30.00&26.96&27.45&27.08&34.28&38.77&25.00&23.11&29.75&30.92&26.64&25.96&30.97\\
        \cmidrule(r){2-17}
        Average&21.11&23.84&21.52&18.24&17.91&21.39&19.36&22.90&25.08&18.36&16.52&18.40&21.04&20.99&18.96&19.46\\
        \bottomrule
        \end{tabular}
    }
    \captionsetup{justification=justified, singlelinecheck=false}
    \caption{
    Impact of different prompting strategies on model performance. The table compares the effectiveness of four prompting methods: \textbf{Historian Role-playing (Role)}, \textbf{Professional Prompting (Prof.)}, \textbf{Chain-of-Thought (CoT)}, and \textbf{Retrieval-Augmented Generation (RAG)} across Term Interpretation, Fact QA, and Historical Reasoning tasks. 
    }
    \label{tab:promptdetail}
\end{table*}

\section{Detailed Evaluation of RAG}~\label{app:rag}
\paragraph{Implementation of RAG.}
We construct the retrieval corpus by cleaning, segmenting, and vectorizing the latest Chinese Wikipedia dump. Our RAG system is implemented with the FlashRAG framework~\cite{DBLP:conf/www/Jin0DDYZZYW25}, using \texttt{bge-large-zh-v1.5}\footnote{\url{https://huggingface.co/BAAI/bge-large-zh-v1.5}} as the embedding model.

In this section, we examine how varying the number of retrieved documents ($k$) affects model performance. As shown in Table~\ref{tab:rag}, increasing $k$ does not lead to consistent gains. Instead, most LLMs exhibit substantial sensitivity to retrieval noise. For example, for Gemini-3-Pro-Preview, the average score decreases from 27.35 to 25.51 as $k$ increases from 10 to 100. This suggests that, for models with relatively strong internal knowledge, retrieving too many external documents (e.g., $k=50$ or $100$) may introduce irrelevant information that interferes with otherwise precise reasoning.

\begin{table*}[h!]
    \centering
    % \small
    \renewcommand{\arraystretch}{1.1}
    \setlength{\tabcolsep}{1.2mm}
    \resizebox{1.0\linewidth}{!}{% 
        \begin{tabular}{l 
        *{4}{S[table-format=2.2, mode=text]}   % Level 1 的4列
      !{\hspace{2mm}}                  % <--- 【关键】Level 1 和 2 之间的大间距
      *{4}{S[table-format=2.2, mode=text]}   % Level 2 的4列
      !{\hspace{2mm}}                  % <--- 【关键】Level 2 和 3 之间的大间距
      *{4}{S[table-format=2.2, mode=text]}   % Level 3 的4列
      !{\hspace{2mm}}                  % <--- 【关键】Level 3 和 Average 之间的大间距
      *{4}{S[table-format=2.2, mode=text]}   % Average 的4列
        }
        \toprule
        \multirow{3}{*}{\textbf{Model}} & 
      \multicolumn{4}{c}{RAG ($k=10$)} & 
      \multicolumn{4}{c}{RAG ($k=20$)} \\
      \cmidrule(lr){2-5} \cmidrule(lr){6-9}
        & {\makecell{Concept \\Interpretation}} & {\makecell{Fact \\Exposition}} & {\makecell{Historical \\Reasoning}}& {Average} 
      & {\makecell{Concept \\Interpretation}} & {\makecell{Fact \\Exposition}} & {\makecell{Historical \\Reasoning}} & {Average} \\
        \midrule
        \multicolumn{9}{c}{\cellcolor[HTML]{EFEFEF}\textit{Closed-Source Models}} \\
        \midrule
        Claude-sonnet-4.5-20250929-Thinking&11.86&15.33&16.53&14.25&12.65&21.82&16.50&16.10\\
        GPT-5.2&14.41&23.46&19.05&18.65&11.81&22.24&18.04&16.35\\
        GPT-5.2-Thinking&16.10&21.54&21.16&19.25&16.39&26.71&19.75&19.94\\
        GPT-o3-2025-04-16&20.28&19.79&24.58&21.25&20.84&27.76&21.42&22.67\\
        Gemini-3-Pro-Preview&25.17&28.83&28.79&27.35&26.48&24.59&26.50&26.04\\
        Gemini-3-Pro-Preview-Thinking&22.62&28.88&32.11&27.21&24.97&24.82&24.83&24.89\\
        Qwen3-Max&17.24&24.17&24.95&21.58&17.94&26.00&24.17&21.92\\
        \midrule
        \multicolumn{9}{c}{\cellcolor[HTML]{EFEFEF}\textit{Open-Source Models}} \\
        \midrule
        Llama-4-Scout-17B-16E-Instruct&4.62&6.54&3.74&5.03&5.00&7.12&7.96&6.49\\
        gpt-oss-120b-mxfp4-high&9.97&12.08&13.42&11.58&11.61&14.35&11.42&12.19\\
        gpt-oss-20b-mxfp4-high&1.72&2.38&0.37&1.58&1.19&2.88&6.96&3.51\\
        Kimi-K2-Thinking&23.62&25.29&24.47&24.40&22.19&28.82&25.79&24.96\\
        GLM-4.6-Thinking&13.69&22.21&20.42&18.31&17.10&22.94&15.63&17.99\\
        Qwen3-14B-Thinking&8.90&14.67&12.00&11.64&9.16&16.82&11.79&11.85\\
        Qwen3-32B-Thinking&11.03&13.17&14.47&12.65&12.81&11.29&14.29&12.94\\
        Qwen3-235B-A22B-Thinking&17.93&18.29&22.16&19.17&16.74&20.94&15.25&17.24\\
        DeepSeek-V3.2&21.07&23.92&22.89&22.50&23.55&24.65&24.75&24.21\\
        DeepSeek-V3.2-Thinking&21.55&24.54&28.53&24.39&17.81&30.24&23.96&22.79\\
        DeepSeek-R1-0528&25.24&24.58&26.00&25.22&20.13&32.35&29.13&26.01\\
        \toprule
        \multirow{3}{*}{\textbf{Model}} & 
      \multicolumn{4}{c}{RAG ($k=50$)} & 
      \multicolumn{4}{c}{RAG ($k=100$)} \\
      \cmidrule(lr){2-5} \cmidrule(lr){6-9}
        & {\makecell{Concept \\Interpretation}} & {\makecell{Fact \\Exposition}} & {\makecell{Historical \\Reasoning}}  & {Average} 
      & {\makecell{Concept \\Interpretation}} & {\makecell{Fact \\Exposition}} & {\makecell{Historical \\Reasoning}}  & {Average} \\
        \midrule
        \multicolumn{9}{c}{\cellcolor[HTML]{EFEFEF}\textit{Closed-Source Models}} \\
        \midrule
        Claude-sonnet-4.5-20250929-Thinking&20.72&12.13&11.77&14.13&16.17&16.11&7.71&12.51\\
        GPT-5.2&24.17&15.13&16.58&18.01&17.96&19.67&13.29&16.38\\
        GPT-5.2-Thinking&27.11&19.83&15.52&19.79&19.52&24.17&18.16&20.10\\
        GPT-o3-2025-04-16&28.17&22.17&18.03&21.89&22.09&32.39&23.03&25.07\\
        Gemini-3-Pro-Preview&27.06&27.87&22.81&25.49&26.09&28.22&23.52&25.51\\
        Gemini-3-Pro-Preview-Thinking&27.28&23.04&20.32&22.93&26.22&24.94&20.94&23.63\\
        Qwen3-Max&26.28&17.65&18.26&20.07&19.65&26.28&19.26&21.14\\
        \midrule
        \multicolumn{9}{c}{\cellcolor[HTML]{EFEFEF}\textit{Open-Source Models}} \\
        \midrule
        Llama-4-Scout-17B-16E-Instruct&9.11&4.65&7.45&6.97&7.39&7.44&6.26&6.92\\
        gpt-oss-120b-mxfp4-high&19.56&11.57&5.58&10.99&12.83&9.67&7.74&9.85\\
        gpt-oss-20b-mxfp4-high&3.78&0.61&1.97&1.99&0.61&1.39&1.58&1.22\\
        Kimi-K2-Thinking&31.67&21.30&24.19&25.14&16.57&26.28&19.77&20.38\\
        GLM-4.6-Thinking&19.00&16.09&21.42&19.11&17.35&19.17&17.00&17.65\\
        Qwen3-14B-Thinking&16.56&7.35&11.19&11.31&7.70&18.22&8.84&10.82\\
        Qwen3-32B-Thinking&18.89&11.78&9.52&12.58&12.57&17.94&10.58&13.06\\
        Qwen3-235B-A22B-Thinking&18.78&13.70&15.74&15.85&16.78&31.89&14.74&19.68\\
        DeepSeek-V3.2&25.44&21.39&20.19&21.89&22.22&24.72&20.19&21.97\\
        DeepSeek-V3.2-Thinking&33.78&18.91&20.16&23.17&22.26&24.00&22.77&22.92\\
        DeepSeek-R1-0528&25.83&19.65&25.16&23.57&18.17&23.94&23.58&21.94\\
        \bottomrule
        \end{tabular}
    }
    \captionsetup{justification=justified, singlelinecheck=false}
    \caption{
    Performance comparison of LLMs with varying numbers of retrieved documents ($k$) provided as context. Here, $k \in \{10, 20, 50, 100\}$ denotes the specific number of top-ranked documents retrieved and concatenated into the input prompt for each model.
    }
    \label{tab:rag}
\end{table*}
\section{Additional Experimental Analysis and Results}~\label{app-moreanalysis}

\subsection{Difficulty-wise Performance Analysis}

\begin{figure*}[h!]
    \centering
    \includegraphics[width=1.0\linewidth]{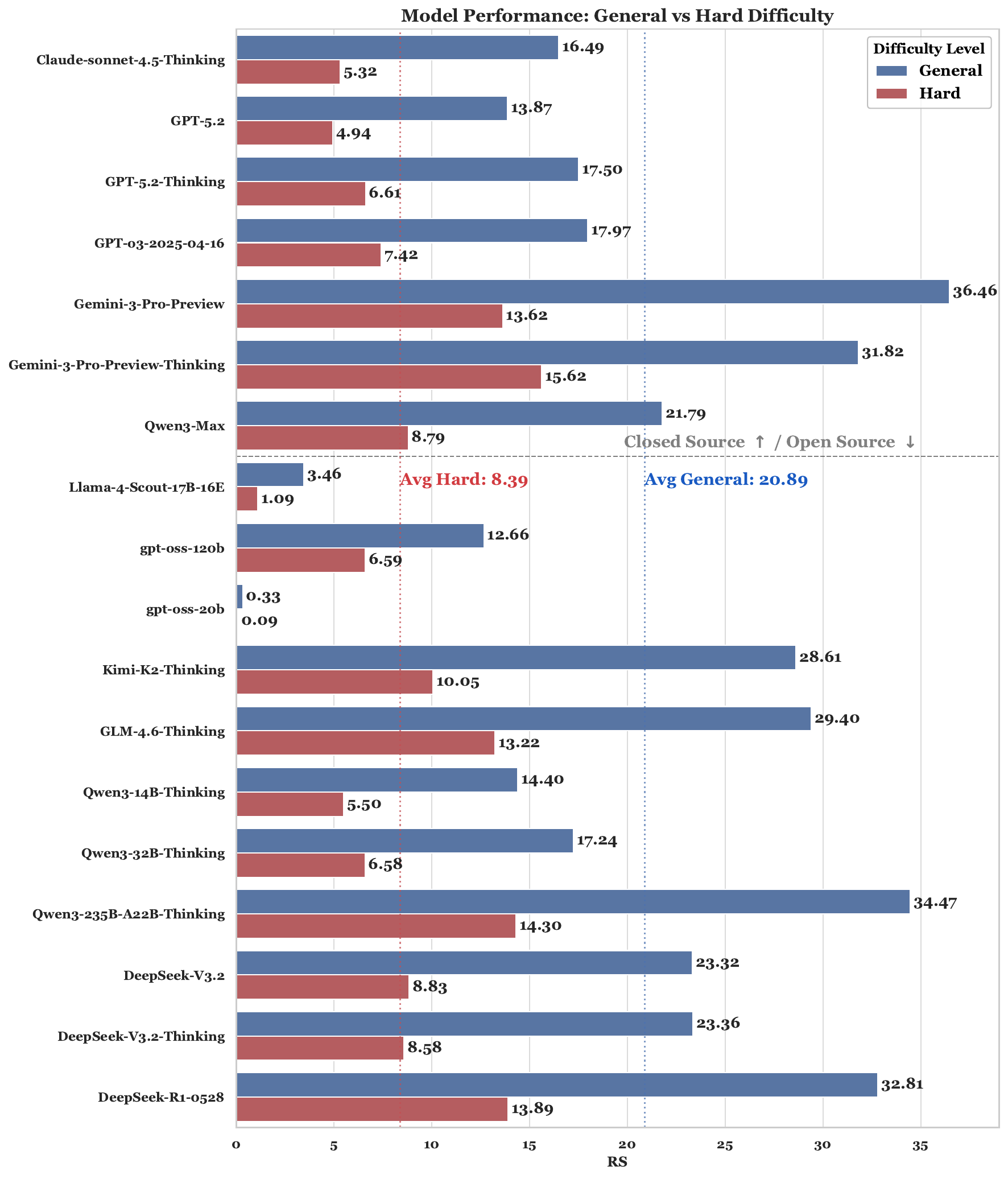}
    \caption{
    Performance gap between General and Hard difficulty levels.
The sharp decline in average scores (20.89 to 8.39) confirms the benchmark's rigor in distinguishing deep historical reasoning from basic factual recall.
    }
    \label{fig:difficulty_res}
\end{figure*}

As illustrated in Figure~\ref{fig:difficulty_res}, the transition from General to Hard tasks precipitates a universal performance decline, with the average score plummeting from 20.89 to 8.39. This $>50\%$ drop confirms that while current LLMs possess broad historical knowledge, they lack the deep inferential capabilities required for complex historiography. Notably, models incorporating chain-of-thought mechanisms—such as Gemini-3-Pro-Preview-Thinking (15.62) and Qwen3-235B-A22B-Thinking—demonstrate superior resilience compared to standard models, indicating that explicit reasoning steps are essential for navigating complex historical logic. 

\subsection{Topic-wise Performance Analysis}
As detailed in Table~\ref{tab:topic-perf}, model performance exhibits sharp variance across historical sub-domains, revealing a distinct dichotomy between static knowledge and complex reasoning. Specifically, while models consistently achieve higher scores in factual tasks like \textit{Keju Literature} and \textit{Ancient Education}, their proficiency drops notably in reasoning-intensive domains such as \textit{Keju Reform} and \textit{Ancient Geo \& Transport}, which demand the synthesis of spatial and temporal changes. This analysis also highlights a significant capability gap: top-tier closed-source models (e.g., Gemini-3-Pro) maintain robustness across diverse topics, whereas open-source models struggle with complex contexts—exemplified by near-zero scores in \textit{Ancient Social Networks} due to their inability to capture implicit interpersonal relations.

\begin{table*}[h!]
    \centering
    % \small
    \renewcommand{\arraystretch}{1.1}
    \setlength{\tabcolsep}{1.2mm}
    \resizebox{1.0\linewidth}{!}{% 
\begin{tabular}{lccccccc}
\toprule
\textbf{Model} & 
\makecell[c]{\textbf{Ancient} \\ \textbf{Social Net}} & 
\makecell[c]{\textbf{Ancient Geo} \\ \textbf{\& Transport}} & 
\makecell[c]{\textbf{Ancient} \\ \textbf{Education}} & 
\makecell[c]{\textbf{Ancient} \\ \textbf{Official System}} & 
\makecell[c]{\textbf{Admission \&} \\ \textbf{Career Path}} & 
\makecell[c]{\textbf{Keju Registration} \\ \textbf{\& Process}} & 
\makecell[c]{\textbf{Ancient} \\ \textbf{Academic Thoughts}} \\
\midrule
\multicolumn{8}{c}{\cellcolor[HTML]{EFEFEF}\textit{\textbf{Closed-Source Models}}} \\
\midrule
Claude-sonnet-4.5-20250929-Thinking & 24.00 & 11.17 & 13.75 & 15.70 & 15.23 & 12.34 & 19.00 \\
GPT-5.2 & 15.00 & 15.28 & 11.30 & 12.07 & 12.88 & 12.07 & 16.75 \\
GPT-5.2-Thinking & 27.80 & 10.11 & 15.00 & 17.26 & 18.85 & 14.18 & 23.00 \\
GPT-o3-2025-04-16 & 22.80 & 18.22 & 15.50 & 18.61 & 17.07 & 13.81 & 21.88 \\
Gemini-3-Pro-Preview & 36.20 & 27.33 & 23.02 & 31.22 & 27.48 & 26.82 & 39.88 \\
Gemini-3-Pro-Preview-Thinking & 25.80 & 29.83 & 26.48 & 29.43 & 28.45 & 25.62 & 36.63 \\
Qwen3-Max & 27.40 & 22.72 & 16.14 & 20.54 & 20.93 & 18.12 & 23.50 \\
\midrule
\multicolumn{8}{c}{\cellcolor[HTML]{EFEFEF}\textit{\textbf{Open-Source Models}}} \\
\midrule
Llama-4-Scout-17B-16E-Instruct & 3.20 & 2.44 & 2.36 & 2.72 & 2.48 & 2.78 & 3.75 \\
gpt-oss-120b & 15.80 & 13.50 & 7.86 & 10.59 & 10.90 & 10.78 & 22.38 \\
gpt-oss-20b & 0.00 & 1.22 & 0.20 & 0.11 & 0.35 & 0.20 & 0.00 \\
Kimi-K2-Thinking & 35.40 & 27.11 & 25.09 & 24.52 & 26.37 & 22.00 & 40.25 \\
GLM-4.6-Thinking & 41.20 & 28.72 & 25.23 & 30.57 & 23.07 & 24.40 & 32.25 \\
Qwen3-14B-Thinking & 20.60 & 8.94 & 13.29 & 14.52 & 13.78 & 10.66 & 14.50 \\
Qwen3-32B-Thinking & 17.20 & 10.94 & 13.70 & 19.87 & 16.47 & 14.66 & 22.50 \\
Qwen3-235B-A22B-Thinking & 52.00 & 30.33 & 30.84 & 32.50 & 29.73 & 26.20 & 40.88 \\
DeepSeek-V3.2 & 23.60 & 18.28 & 19.23 & 21.72 & 21.58 & 18.54 & 28.00 \\
DeepSeek-V3.2-Thinking & 22.60 & 22.56 & 18.34 & 23.35 & 23.92 & 18.82 & 22.50 \\
DeepSeek-R1-0528 & 33.80 & 30.44 & 27.13 & 32.52 & 29.67 & 26.84 & 44.38 \\
% \bottomrule
% \end{tabular}

% \vspace{0.5cm} % 两个表格之间的垂直间距
% \begin{tabular}{lcccccc}
\toprule
\textbf{Model} & 
\makecell[c]{\textbf{Ancient Society} \\ \textbf{\& Economy}} & 
\makecell[c]{\textbf{Keju in} \\ \textbf{Human History}} & 
\makecell[c]{\textbf{Keju Reform} \\ \textbf{\& Abolition}} & 
\makecell[c]{\textbf{Keju} \\ \textbf{Literature}} & 
\makecell[c]{\textbf{Keju Definition} \\ \textbf{\& Evolution}} & 
\makecell[c]{\textbf{Keju Materials} \\ \textbf{\& Life Sites}} & /\\
\midrule
\multicolumn{8}{c}{\cellcolor[HTML]{EFEFEF}\textit{\textbf{Closed-Source Models}}} \\
\midrule
Claude-sonnet-4.5-20250929-Thinking & 11.09 & 15.13 & 12.70 & 16.27 & 12.05 & 11.75 &/\\
GPT-5.2 & 8.88 & 3.60 & 10.74 & 12.18 & 10.92 & 13.00 &/\\
GPT-5.2-Thinking & 15.76 & 13.07 & 11.07 & 14.73 & 14.35 & 15.00 &/\\
GPT-o3-2025-04-16 & 13.71 & 19.40 & 13.04 & 20.00 & 13.45 & 15.13 &/\\
Gemini-3-Pro-Preview & 26.65 & 26.20 & 25.30 & 33.19 & 23.52 & 30.38 &/\\
Gemini-3-Pro-Preview-Thinking & 30.06 & 32.47 & 23.11 & 31.09 & 23.81 & 22.13 &/\\
Qwen3-Max & 16.85 & 10.33 & 15.63 & 21.91 & 17.43 & 17.50 &/\\
\midrule
\multicolumn{8}{c}{\cellcolor[HTML]{EFEFEF}\textit{\textbf{Open-Source Models}}} \\
\midrule
Llama-4-Scout-17B-16E-Instruct & 2.09 & 1.13 & 1.67 & 6.55 & 2.12 & 4.38 &/\\
gpt-oss-120b & 11.88 & 12.60 & 9.93 & 8.18 & 10.23 & 14.00 &/\\
gpt-oss-20b & 0.91 & 0.00 & 0.44 & 0.00 & 0.20 & 0.00 &/\\
Kimi-K2-Thinking & 24.68 & 13.80 & 17.44 & 26.36 & 21.40 & 28.13 &/\\
GLM-4.6-Thinking & 20.59 & 26.13 & 24.26 & 29.64 & 22.18 & 26.25 &/\\
Qwen3-14B-Thinking & 11.26 & 9.87 & 8.52 & 15.64 & 10.49 & 20.50 &/\\
Qwen3-32B-Thinking & 13.41 & 7.13 & 11.81 & 16.64 & 14.42 & 16.12 \\
Qwen3-235B-A22B-Thinking & 26.59 & 23.20 & 26.48 & 27.27 & 27.17 & 32.75 &/\\
DeepSeek-V3.2 & 19.47 & 13.33 & 12.93 & 24.91 & 15.82 & 20.00&/\\
DeepSeek-V3.2-Thinking & 16.71 & 13.93 & 12.37 & 25.36 & 14.60 & 24.13 &/\\
DeepSeek-R1-0528 & 24.12 & 23.00 & 22.11 & 29.36 & 25.68 & 25.75 &/\\
\bottomrule
\end{tabular}
    }
    \captionsetup{justification=justified, singlelinecheck=false}
    \caption{
    Fine-grained performance across specific historical topics. The table reports the evaluation scores of varying LLMs on 13 distinct topics. 
    }
    \label{tab:topic-perf}
\end{table*}

\subsection{Dynasty-Level Performance Analysis}
\begin{figure}[t]
    \centering
    \includegraphics[width=1.0\linewidth]{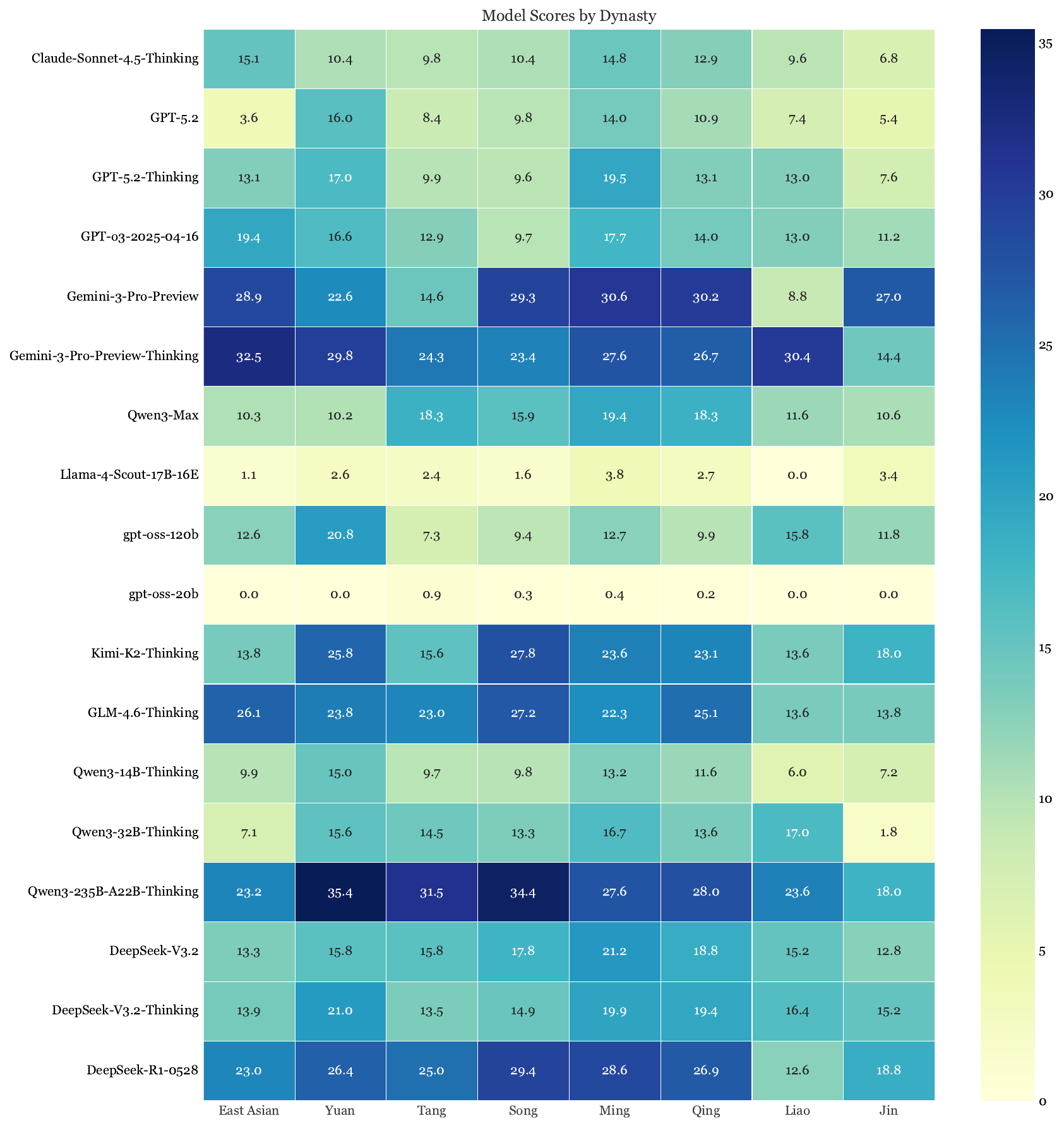}
    \caption{
    Heatmap of LLM performance across diverse historical periods. Most LLMs perform robustly in mainstream eras (e.g., Tang, Song, Ming) but struggle in low-resource contexts like the Liao and Jin dynasties.
    }
    \label{fig:Dynasty_res}
\end{figure}

Figure~\ref{fig:Dynasty_res} reveals a marked imbalance in model proficiency across historical eras, correlating strongly with textual data availability. While top-tier LLMs (e.g., Qwen3-235B-Thinking, Gemini-3-Pro) demonstrate robustness in mainstream dynasties such as Tang, Song, and Ming—achieving peak RS exceeding 30.0—performance universally degrades in low-resource contexts like the Liao and Jin dynasties, where scores often drop significantly. This result highlight that current LLMs struggle to transfer reasoning capabilities to historical periods with sparser textual records.

\subsection{Detailed Results of Different LLMs on the T1-T3 Tasks}~\label{app-maint13}

Detailed results of different LLMs on the T1-T3 tasks are reported in this section. Table~\ref{tab:appmain} presents the performance of representative
closed-source and open-source LLMs

\begin{table*}[t]
    \centering
    \large
    \renewcommand{\arraystretch}{1}
    \resizebox{1.0\linewidth}{!}{% 
        \begin{tabular}{lcccccccccccccccc
        }
        \toprule
        \multirow{2}{*}{\textbf{Model}} & 
    \multicolumn{4}{c}{\textbf{Term Interpretation}} & \multicolumn{4}{c}{\textbf{Fact QA}}  & \multicolumn{4}{c}{\textbf{Historical Reasoning}}  & 
    \multicolumn{4}{c}{\textbf{Average}} \\
    \cmidrule(lr){2-5}\cmidrule(lr){6-9}\cmidrule(lr){10-13}\cmidrule(lr){14-17}
        % \cmidrule(lr){5-8}
         &{BL} & {RG} & {BS} & {RS} & {BL} & {RG} & {BS} & {RS} & {BL} & {RG} & {BS} & {RS} & {BL} & {RG} & {BS} & {RS}\\
        \midrule
        \multicolumn{17}{c}{\cellcolor[HTML]{EFEFEF}\textit{Closed-Source Models}} \\
        \midrule
        Claude-Sonnet-4.5-Thinking&1.42&1.76&71.24&12.72&3.36&5.12&71.68&15.21&2.33&6.57&71.42&11.87&2.53&4.76&71.49&12.99\\
        GPT-5.2&2.39&0.32&70.12&11.55&6.01&4.83&72.05&7.68&1.18&4.09&71.78&12.72&3.50&3.46&71.48&11.07\\
        GPT-5.2-Thinking&2.13&1.06&70.37&15.08&7.96&6.24&71.95&11.31&1.81&5.16&71.95&14.95&4.45&4.58&71.55&14.08\\
        GPT-o3-2025-04-16&2.92&1.12&72.15&14.16&14.95&4.68&72.87&15.10&2.77&4.25&72.59&14.79&7.88&3.65&72.60&14.66\\
        Gemini-3-Pro-Preview&0.60&3.24&73.43&29.57&2.61&6.66&74.15&31.86&2.08&7.98&74.13&27.53&1.94&6.27&73.97&29.29\\
        Gemini-3-Pro-Preview-Thinking&0.63&1.65&73.35&24.42&3.70&6.54&74.18&30.47&1.94&6.08&74.03&26.35&2.35&5.16&73.92&26.73\\
        Qwen3-Max&0.00&3.54&74.79&16.35&7.64&7.44&75.14&19.29&4.78&7.96&75.00&17.85&4.77&6.64&75.01&17.71\\
        \midrule
        \multicolumn{17}{c}{\cellcolor[HTML]{EFEFEF}\textit{Open-Source Models}} \\
        \midrule
        Llama-4-Scout-17B-16E&0.00&0.43&71.28&2.51&6.21&3.71&73.11&3.46&0.00&4.29&73.18&2.44&2.59&3.09&72.68&2.72\\
        gpt-oss-120b&0.32&0.22&68.69&11.21&2.27&2.31&70.54&7.96&0.73&2.29&70.85&12.07&1.27&1.78&70.18&10.75\\
        gpt-oss-20b&\textit{Fail}&\textit{Fail}&\textit{Fail}&\textit{Fail}&\textit{Fail}&\textit{Fail}&\textit{Fail}&\textit{Fail}&\textit{Fail}&\textit{Fail}&\textit{Fail}&\textit{Fail}&\textit{Fail}&\textit{Fail}&\textit{Fail}&\textit{Fail}\\
        Kimi-K2-Thinking&2.16&3.44&73.25&22.69&6.10&7.21&73.12&22.67&1.61&7.69&73.28&22.93&3.62&6.43&73.20&22.79\\
        GLM-4.6-Thinking&0.48&1.67&71.26&25.93&3.40&5.88&72.69&24.21&1.65&6.73&72.60&23.10&2.09&5.11&72.30&24.32\\
        Qwen3-14B-Thinking&0.34&0.52&71.87&11.08&3.11&4.87&73.29&13.16&1.53&4.34&73.28&11.09&1.89&3.61&72.93&11.61\\
        Qwen3-32B-Thinking&0.43&0.88&71.97&13.49&3.60&5.16&73.25&13.87&1.35&4.81&73.43&14.23&2.06&3.97&72.99&13.89\\
        Qwen3-235B-A22B-Thinking&0.30&3.83&71.70&26.73&1.62&6.23&72.81&30.36&0.99&5.01&72.70&27.95&1.08&5.22&72.50&28.14\\
        DeepSeek-V3.2&0.93&2.64&72.90&16.56&9.25&7.34&73.68&23.30&1.70&7.44&73.80&17.83&4.65&6.20&73.52&18.77\\
        DeepSeek-V3.2-Thinking&0.68&2.57&73.00&16.29&7.99&7.73&73.48&24.47&4.22&6.88&73.62&17.21&4.91&6.16&73.41&18.72\\
        DeepSeek-R1-0528&0.47&2.83&72.22&25.20&3.45&8.71&73.49&31.31&1.14&6.80&73.42&25.55&1.93&6.60&73.15&26.87\\
        \bottomrule
        \end{tabular}
    }
    \captionsetup{justification=justified, singlelinecheck=false}
    \caption{
    Detailed results of different LLMs on the T1-T3 tasks. The table presents the performance of representative closed-source and open-source LLMs. The metrics reported are \textbf{BLEU (BL)}, \textbf{ROUGE (RG)}, \textbf{BERTScore (BS)}, and \textbf{Rubric Score (RS)}.
    }
    \label{tab:appmain}
\end{table*}

\subsection{Detailed Comparison between Models and Human Performance}
As presented in Table~\ref{tab:app-humanllm}, top-tier LLMs (e.g., Gemini-3-Pro-Preview, Qwen3-235B-Thinking) effectively surpass the \textit{Closed-book Historian} baseline (Avg $\sim$34.0 vs. 29.33), demonstrating superior large-scale knowledge retention. However, a substantial gap persists when compared to the \textit{Open-book Historian} (60.67), highlighting that while models excel at static retrieval, they significantly lag behind human experts in complex evidence synthesis and research-oriented reasoning. 

\begin{table}[h!]
    \centering
    % \small
    \renewcommand{\arraystretch}{1.1}
    \setlength{\tabcolsep}{1mm}
    \resizebox{1.0\linewidth}{!}{% 
        \begin{tabular}{lcccc}
        \toprule
\textbf{Model} & \textbf{T1} &\textbf{T2}& \textbf{T3}& \textbf{Average} \\
\midrule
\multicolumn{5}{c}{\cellcolor[HTML]{EFEFEF}\textit{Closed-Source Models}} \\
\midrule
Claude-Sonnet-4.5-Thinking & 21.40 & 15.50 & 21.10 & 19.33 \\
GPT-5.2 & 6.40 & 21.90 & 16.20 & 14.83 \\
GPT-5.2-Thinking & 13.40 & 14.30 & 18.20 & 15.30 \\
GPT-o3-2025-04-16 & 21.40 & 17.20 & 19.50 & 19.37 \\
Gemini-3-Pro-Preview & 35.60 & 39.60 & 27.30 & 34.17 \\
Gemini-3-Pro-Preview-Thinking & 36.90 & 37.90 & 24.90 & \textbf{33.23} \\
Qwen3-Max & 28.80 & 28.90 & 23.10 & 26.93 \\
\midrule
\multicolumn{5}{c}{\cellcolor[HTML]{EFEFEF}\textit{Open-Source Models}}\\
\midrule
Llama-4-Scout-17B-16E-Instruct & 4.40 & 5.00 & 3.80 & 4.40 \\
gpt-oss-120b & 0.00 & 7.10 & 12.90 & 6.67 \\
gpt-oss-20b & 5.00 & 0.00 & 0.00 & 1.67 \\
Kimi-K2-Thinking & 29.20 & 32.00 & 24.20 & 28.47 \\
GLM-4.6-Thinking & 26.20 & 36.70 & 20.30 & 27.73 \\
Qwen3-14B-Thinking & 16.50 & 14.90 & 15.40 & 15.60 \\
Qwen3-32B-Thinking & 22.60 & 20.50 & 30.20 & 24.43 \\
Qwen3-235B-A22B-Thinking & 31.20 & 36.90 & 33.80 & \textbf{33.97} \\
DeepSeek-V3.2 & 27.80 & 23.60 & 16.90 & 22.77 \\
DeepSeek-V3.2-Thinking & 25.20 & 19.00 & 17.60 & 20.60 \\
DeepSeek-R1-0528 & 33.00 & 36.60 & 24.20 & 31.27 \\
\midrule
\multicolumn{5}{c}{\cellcolor[HTML]{EFEFEF}\textit{Human}}\\
\midrule
Historian (Closed-book) &35.00&22.00&31.00&29.33 \\
Historian (Open-book) &74.00&52.00&56.00&60.67 \\
\bottomrule
\end{tabular}
    }
    \captionsetup{justification=justified, singlelinecheck=false}
    \caption{
    Performance comparison between SOTA LLMs and human experts. The evaluation contrasts model performance against \textit{Closed-book} (memory-based) and \textit{Open-book} (research-based) human baselines across three task dimensions (T1-T3).
    }
    \label{tab:app-humanllm}
\end{table}

\begin{table*}[h!]
    \centering
    % \small
    \renewcommand{\arraystretch}{1.1}
    \setlength{\tabcolsep}{1.2mm}
    \resizebox{1.0\linewidth}{!}{% 
        \begin{tabular}{lccccc
        }
        \toprule
        Model&Open-Source&\# Params&Institution&Deployment&Domain\\
        \midrule
Claude-Sonnet-4.5-Thinking~\cite{anthropic2025claude-sonnet-4-5}&No&-&Anthropic&Official API&General\\
GPT-5.2~\cite{openai2025gpt5-2}&No&-&OpenAI&Official API&General\\
GPT-5.2-Thinking~\cite{openai2025gpt5-2}&No&-&OpenAI&Official API&General\\
GPT-o3~\cite{openai2025o3}&No&-&OpenAI&Official API&General\\
Gemini-3-Pro-Preview~\cite{deepmind2025gemini3pro}&No&-&Google DeepMind&Official API&General\\
Gemini-3-Pro-Preview-Thinking~\cite{deepmind2025gemini3pro}&No&-&Google DeepMind&Official API&General\\
Qwen3-Max~\cite{qwen2025qwen3max}&No&> 1T&Alibaba&Official API&General\\
Llama-4-Scout-17B-16E~\cite{meta2025llama4}&Yes&109B (17B active)&Meta&Locally Load&General\\
gpt-oss-120b~\cite{openai2025gptoss120bgptoss20bmodel}&Yes&116.8B (5.1B active)&OpenAI&Locally Load&General\\
gpt-oss-20b~\cite{openai2025gptoss120bgptoss20bmodel}&Yes&20.9B (3.6B active)&OpenAI&Locally Load&General\\
Kimi-K2-Thinking~\cite{kimiteam2025kimik2openagentic}&Yes&1T (32B active)&Moonshot AI&Locally Load&General\\
GLM-4.6-Thinking~\cite{5team2025glm45agenticreasoningcoding}&Yes&335B (32B active)&Zhipu AI&Locally Load&General\\
Qwen3-14B-Thinking~\cite{yang2025qwen3}&Yes&14B&Alibaba&Locally Load&General\\
Qwen3-32B-Thinking~\cite{yang2025qwen3}&Yes&32B&Alibaba&Locally Load&General\\
Qwen3-235B-A22B-Thinking~\cite{yang2025qwen3}&Yes&235B (22B active)&Alibaba&Locally Load&General\\
DeepSeek-V3.2~\cite{deepseekai2025deepseekv32pushingfrontieropen}&Yes&-&DeepSeek-AI&Locally Load&General\\
DeepSeek-V3.2-Thinking~\cite{deepseekai2025deepseekv32pushingfrontieropen}&Yes&-&DeepSeek-AI&Locally Load&General\\
DeepSeek-R1-0528~\cite{guo2025deepseekr1}&Yes&671B (37B active)&DeepSeek-AI&Locally Load&General\\
        \bottomrule
        \end{tabular}
    }
    % \captionsetup{justification=justified, singlelinecheck=false}
    \caption{
   Details of all evaluated LLMs. 
    }
    \label{tab:llmdetail}
\end{table*}

\section{Case Study}~\label{app-case}
\begin{figure*}[p]
    \centering
    % \begin{subfigure}{\linewidth}
        \centering
        \includegraphics[width=\linewidth]{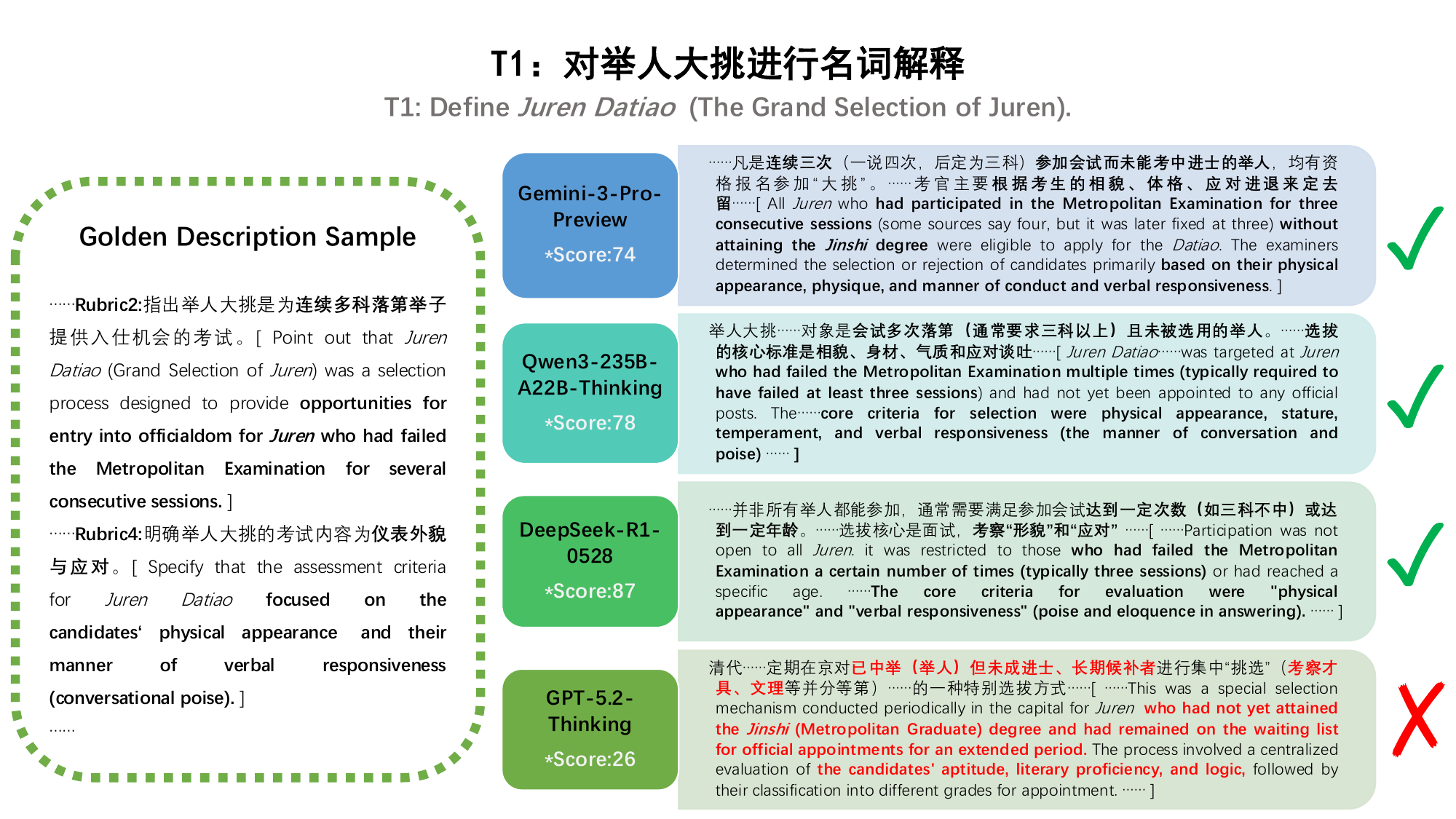}
    % \end{subfigure}
    % \begin{subfigure}{\linewidth}
        \centering
        \includegraphics[width=\linewidth]{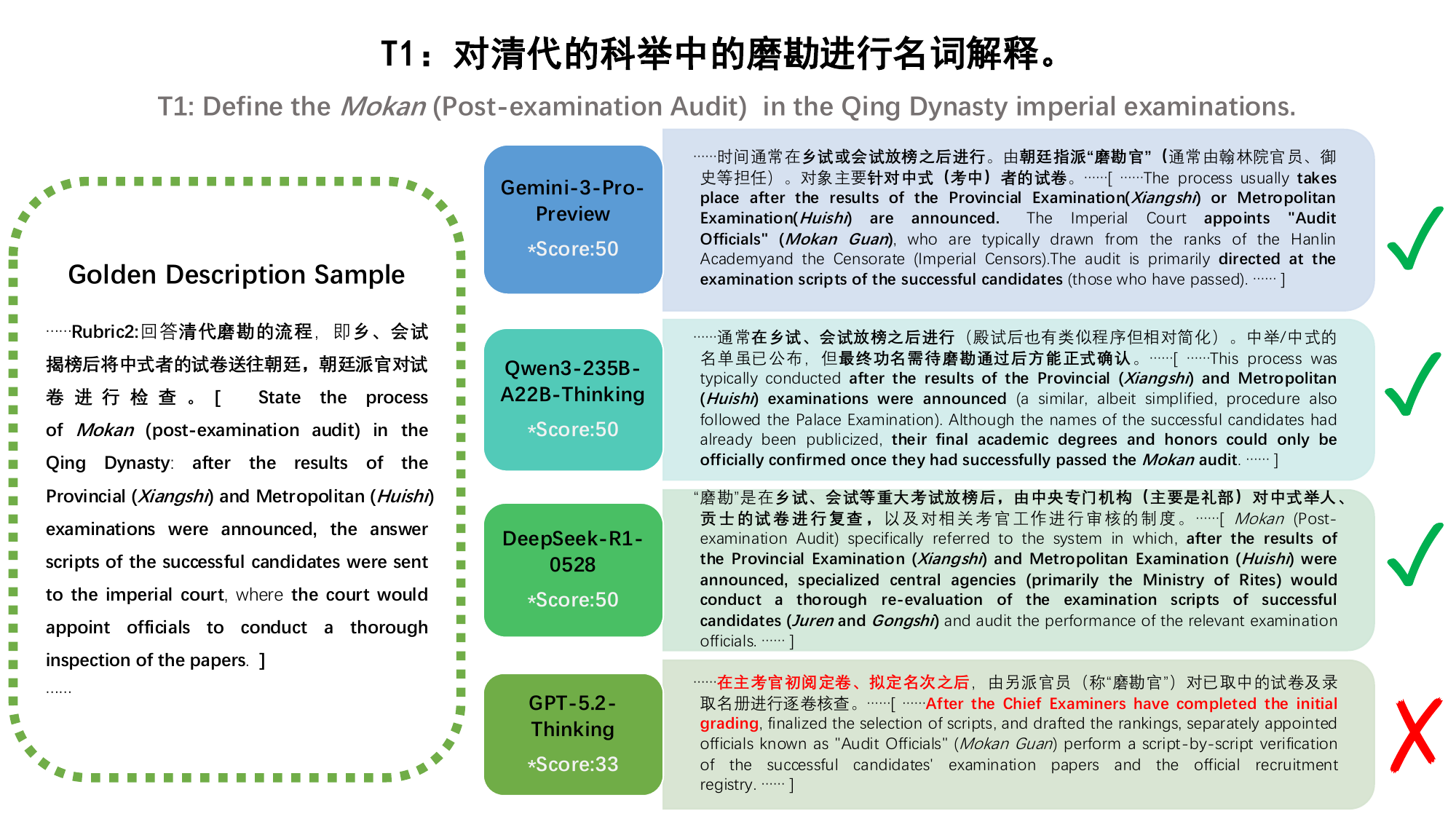}
    % \end{subfigure}
    \caption{Case study on Task T1 (Part I): Qualitative Results.}
    \label{fig:full_case_study1}
\end{figure*}

\begin{figure*}[p]
    \centering
    % \begin{subfigure}{\linewidth}
        \centering
        \includegraphics[width=\linewidth]{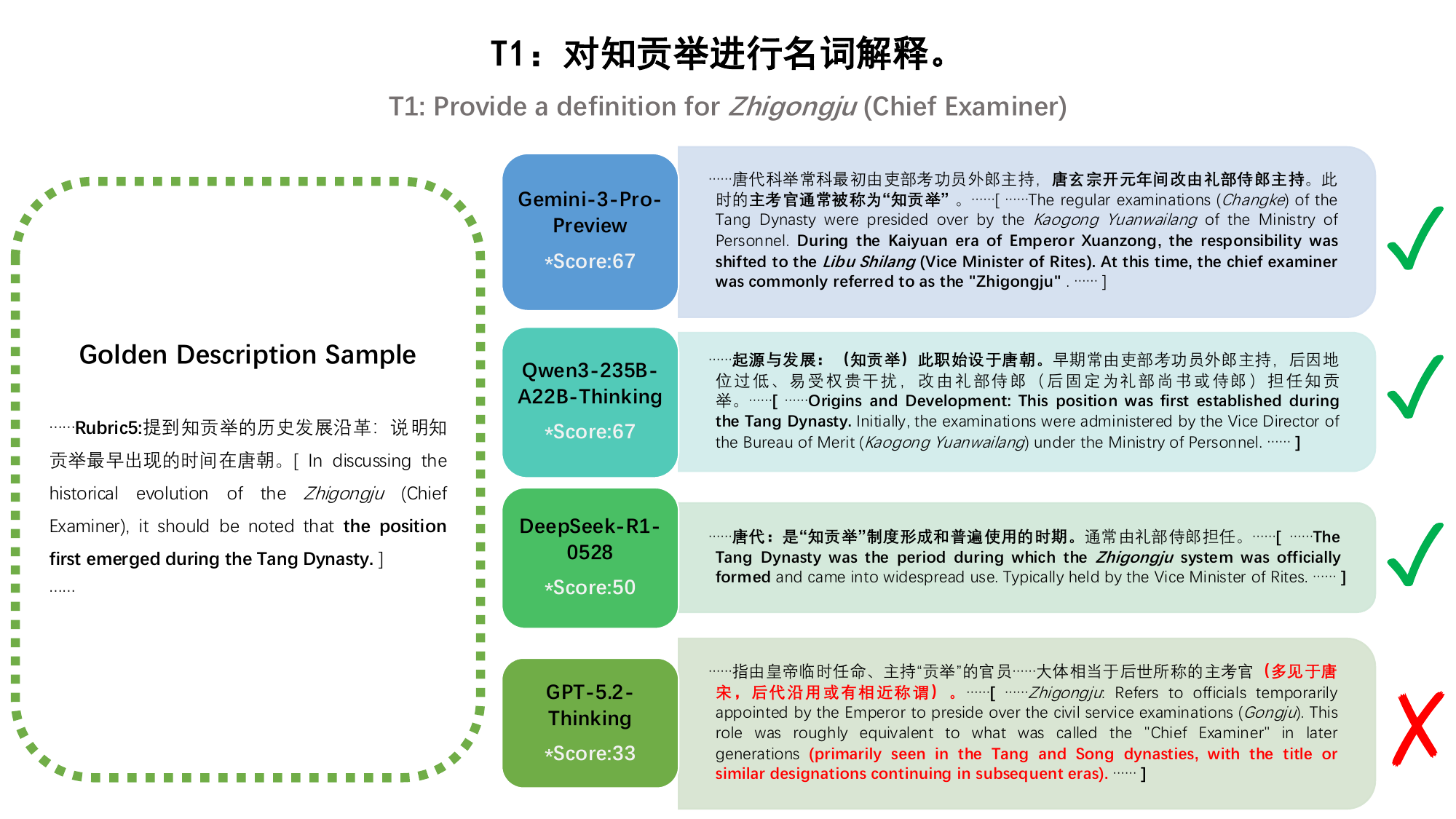}
    % \end{subfigure}
    % \begin{subfigure}{\linewidth}
        \centering
        \includegraphics[width=\linewidth]{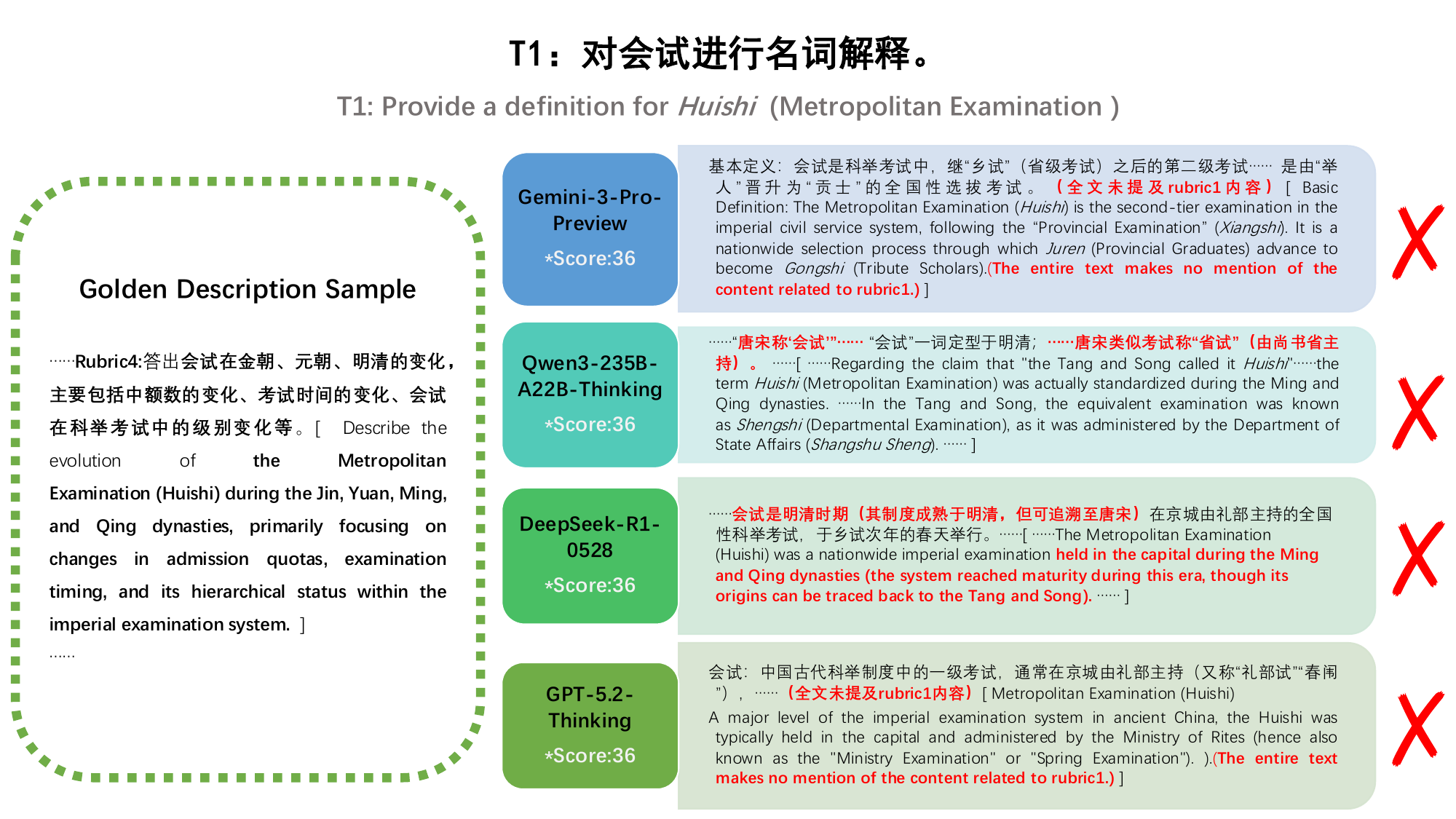}
    % \end{subfigure}
    \caption{Case study on Task T1 (Part II): Qualitative Results.}
    \label{fig:full_case_study2}
\end{figure*}

\begin{figure*}[p]
    \centering
    % \begin{subfigure}{\linewidth}
        \centering
        \includegraphics[width=\linewidth]{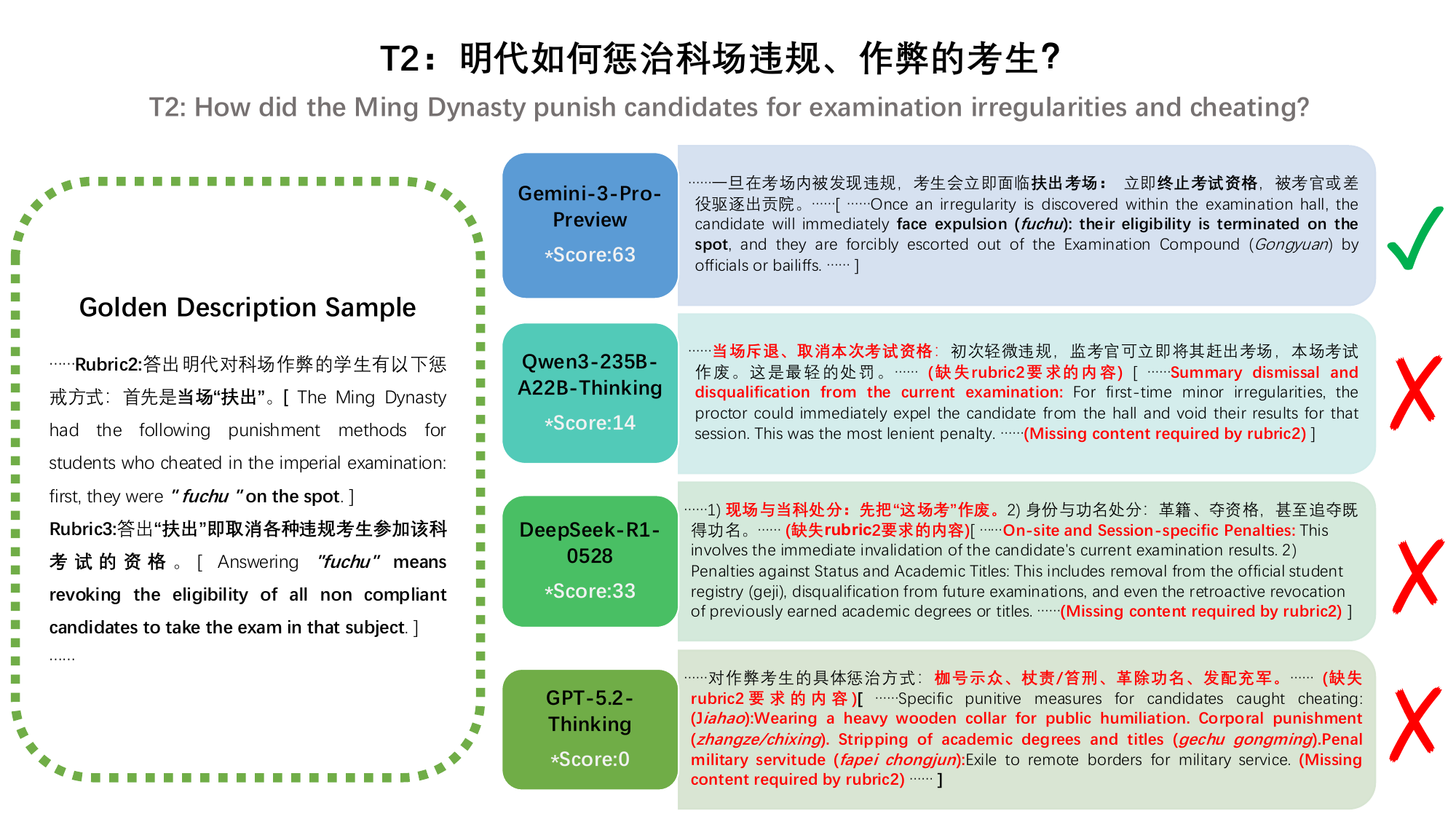}
    % \end{subfigure}
    % \begin{subfigure}{\linewidth}
        \centering
        \includegraphics[width=\linewidth]{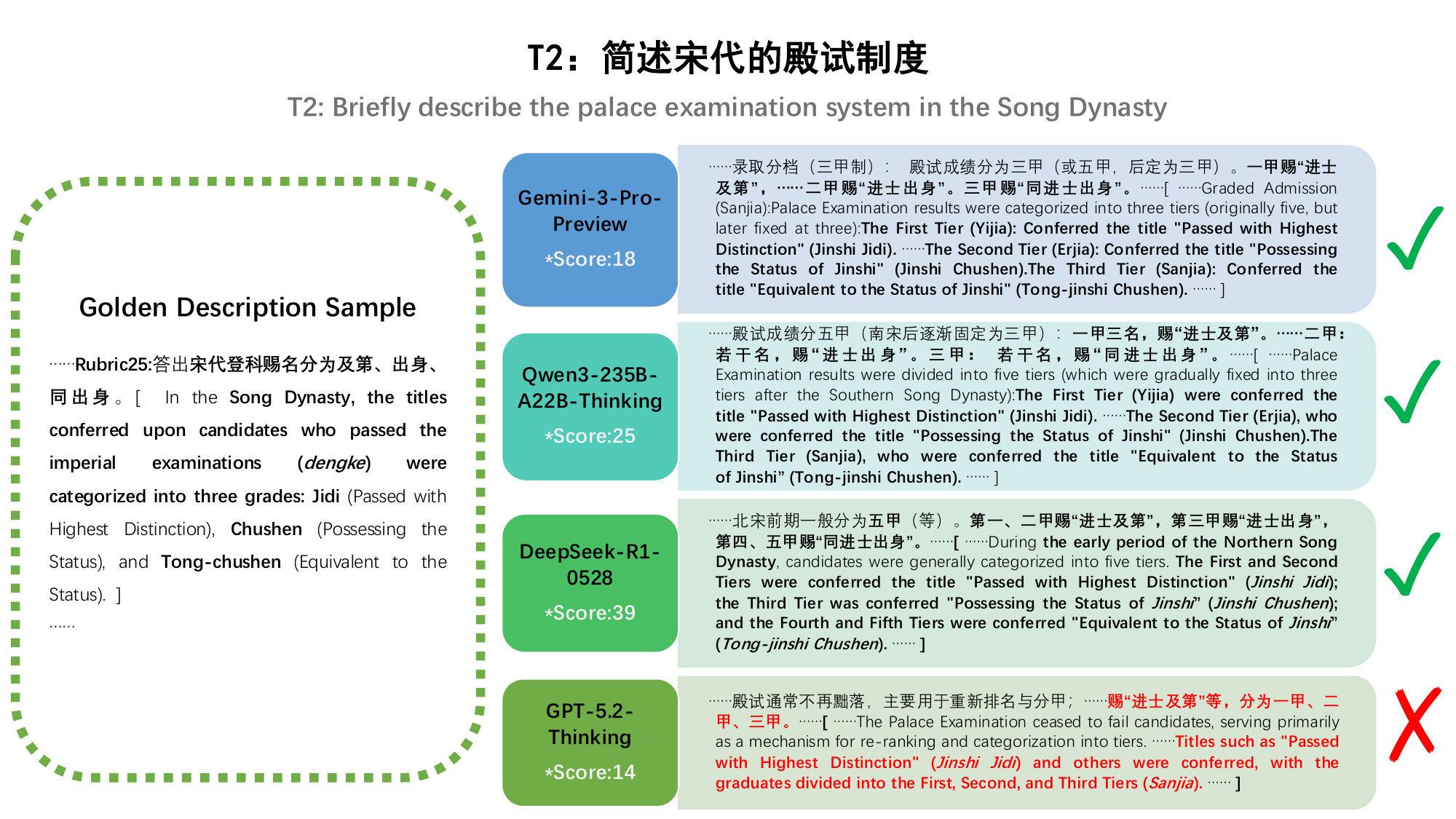}
    % \end{subfigure}
    \caption{Case study on Task T2 (Part I): Qualitative Results.}
    \label{fig:full_case_study3}
\end{figure*}

\begin{figure*}[p]
    \centering
    % \begin{subfigure}{\linewidth}
        \centering
        \includegraphics[width=\linewidth]{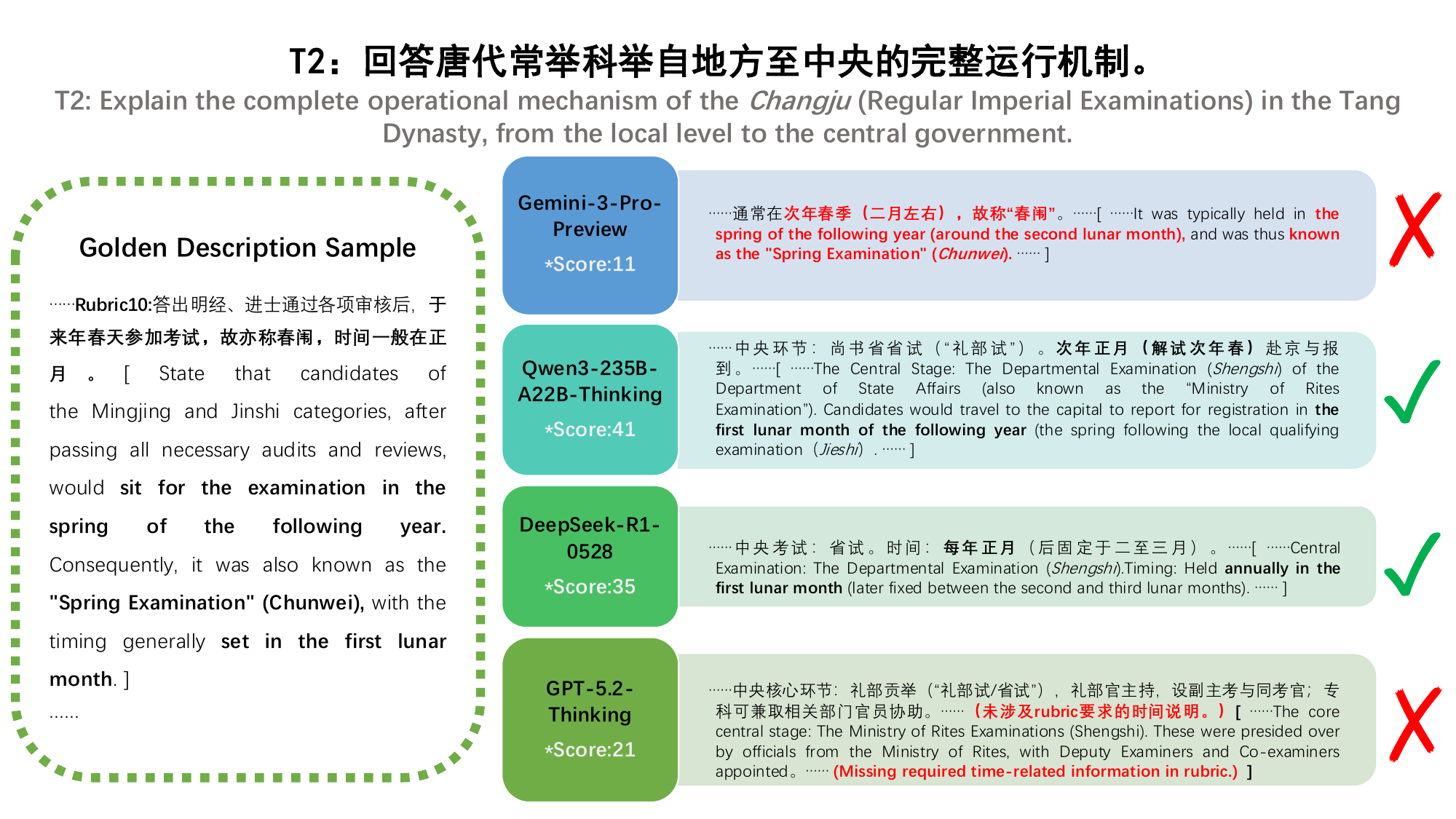}
    % \end{subfigure}
    % \begin{subfigure}{\linewidth}
        \centering
        \includegraphics[width=\linewidth]{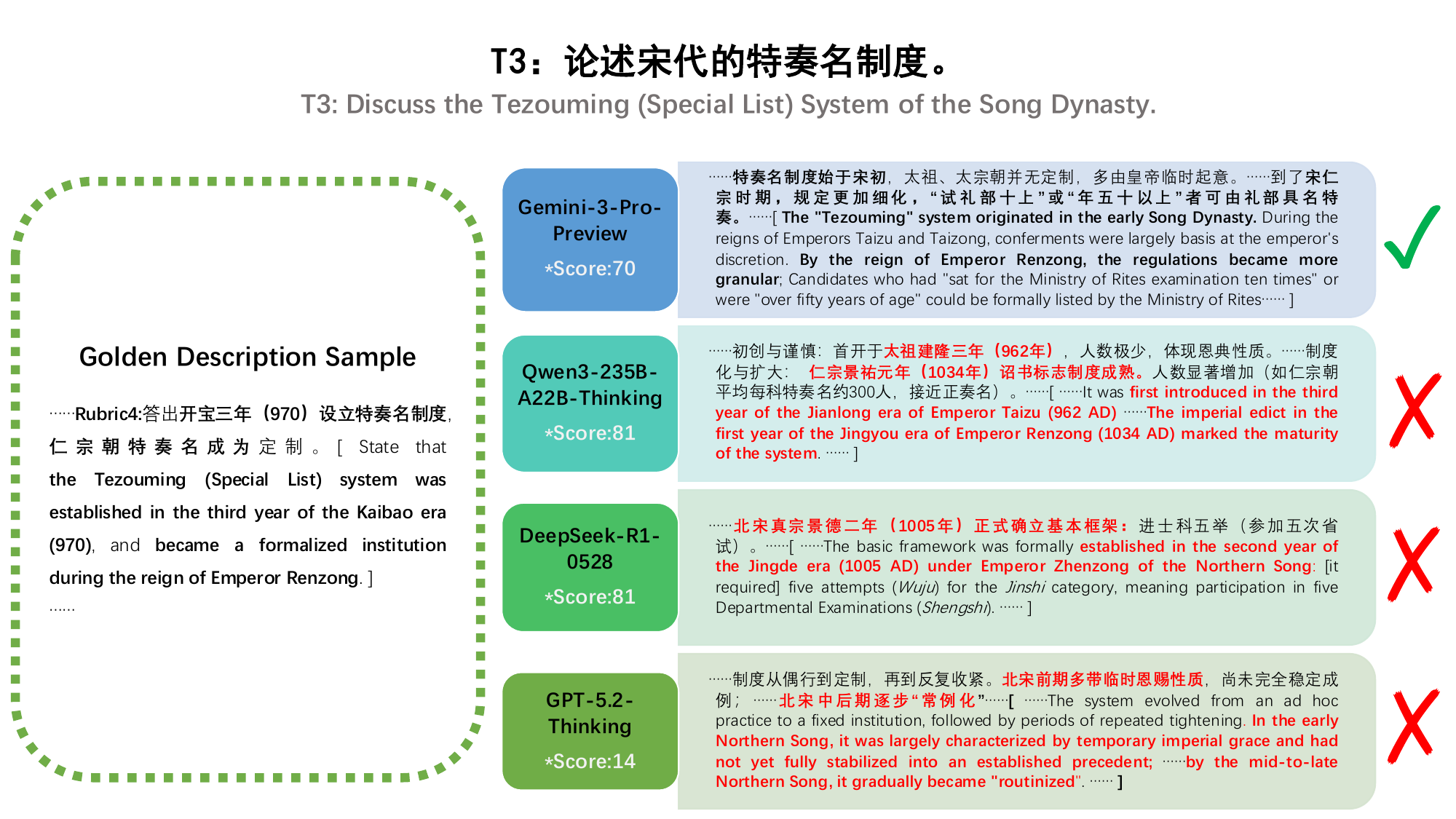}
    % \end{subfigure}
    \caption{Case study on Task T2 (Part II) and T3 (Part I): Qualitative Results.}
    \label{fig:full_case_study4}
\end{figure*}

\begin{figure*}[p]
    \centering
    % \begin{subfigure}{\linewidth}
        \centering
        \includegraphics[width=\linewidth]{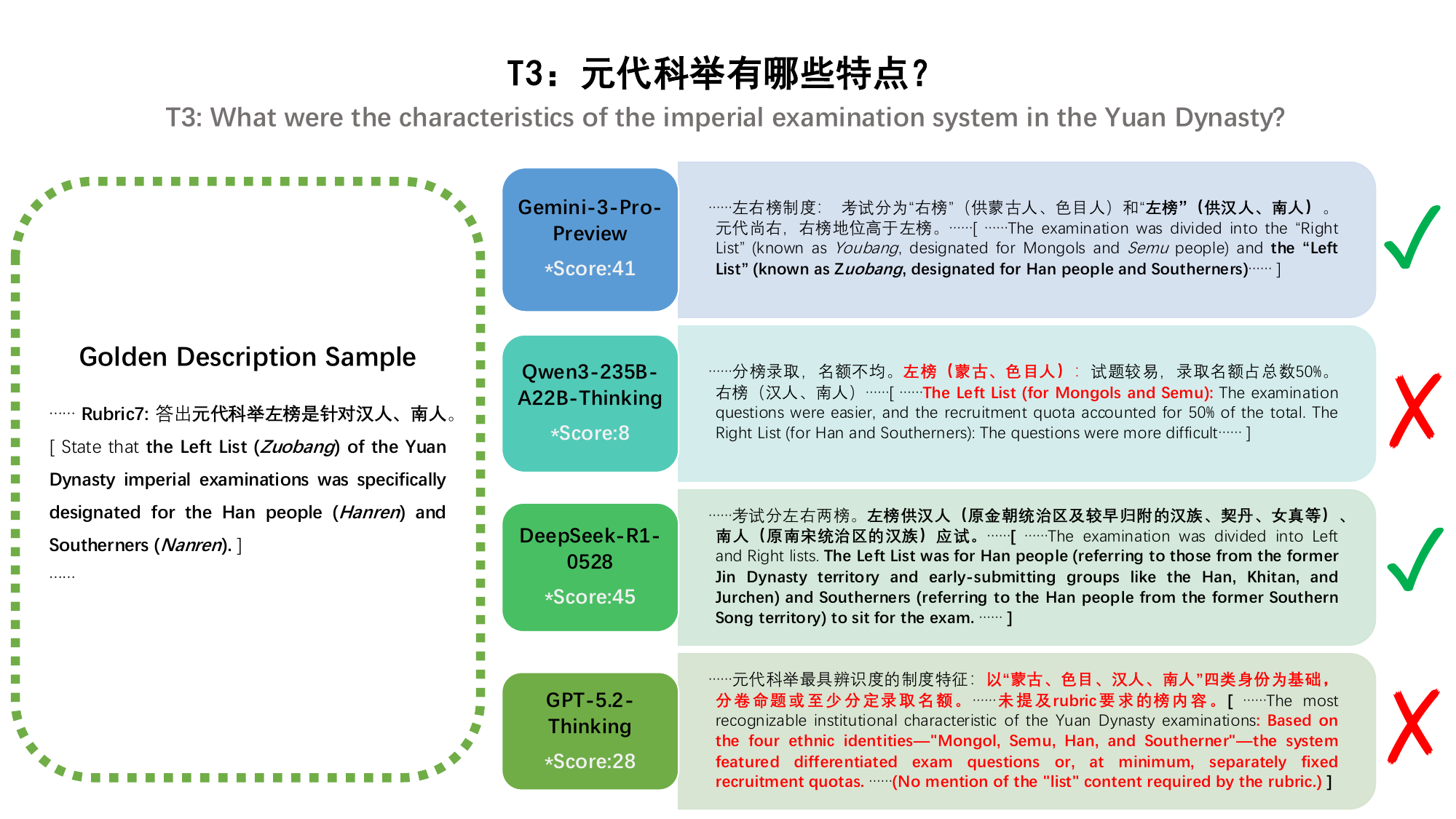}
    % \end{subfigure}
    % \begin{subfigure}{\linewidth}
        \centering
        \includegraphics[width=\linewidth]{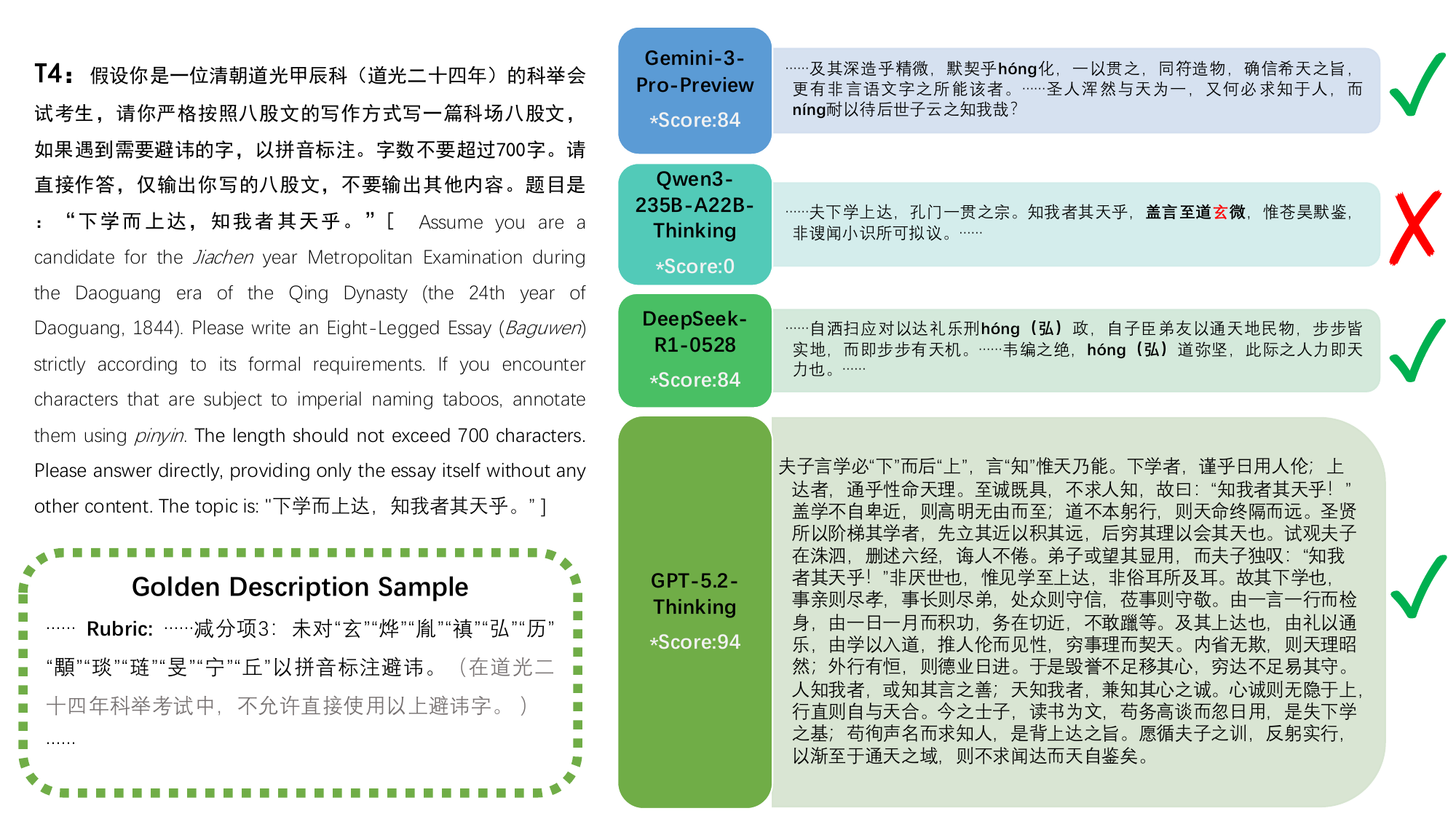}
    % \end{subfigure}
    \caption{Case study on Task T3 (Part II) and T4 (Part I): Qualitative Results.}
    \label{fig:full_case_study5}
\end{figure*}

\begin{figure*}[p]
    \centering
    % \begin{subfigure}{\linewidth}
        \centering
        \includegraphics[width=\linewidth]{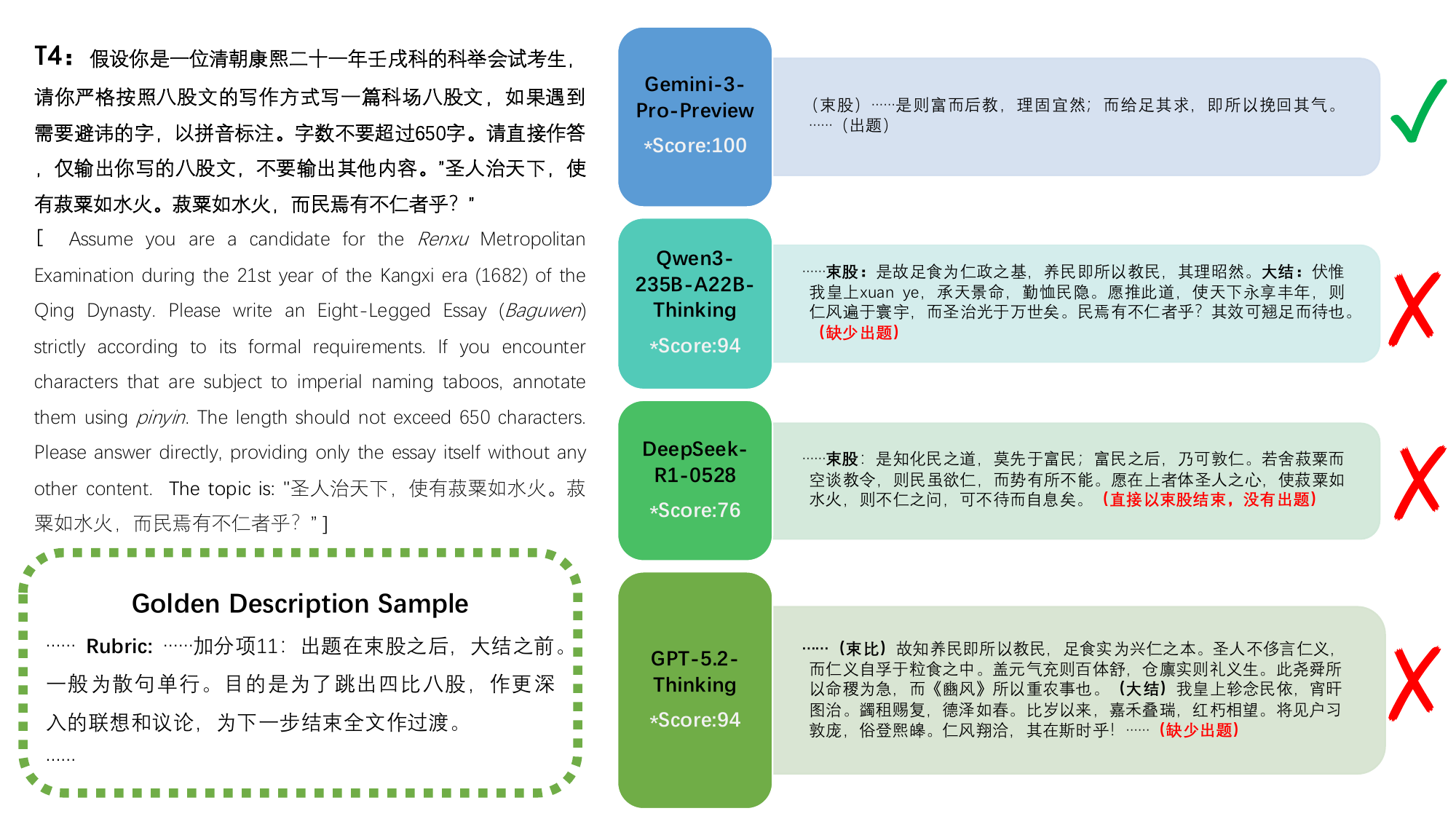}
    % \end{subfigure}
    % \begin{subfigure}{\linewidth}
        \centering
        \includegraphics[width=\linewidth]{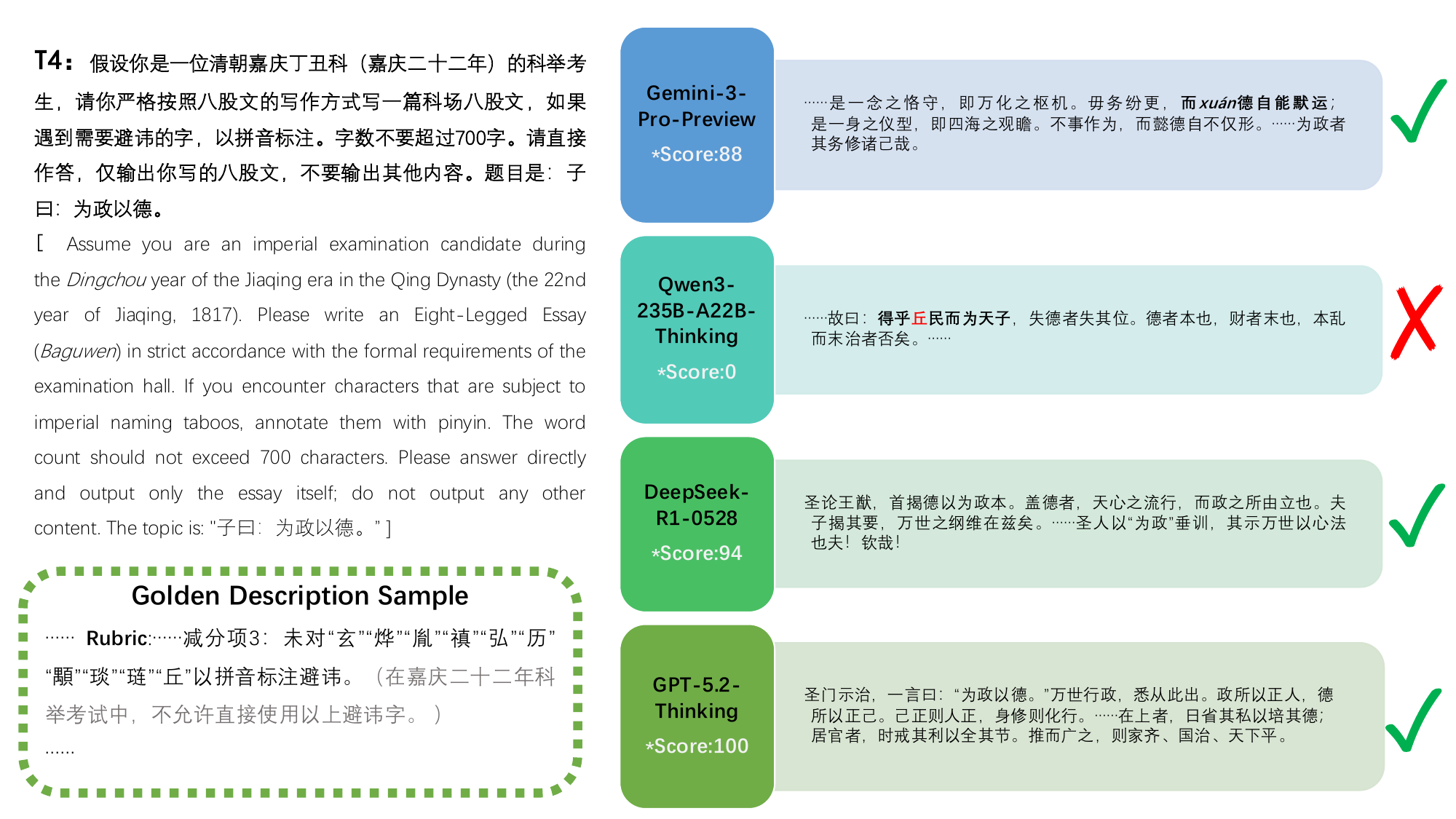}
    % \end{subfigure}
    \caption{Case study on Task T4 (Part II): Qualitative Results.}
    \label{fig:full_case_study6}
\end{figure*}

We conducted detailed case studies on Tasks T1-T4, as shown in Figures~\ref{fig:full_case_study1}, ~\ref{fig:full_case_study2},~\ref{fig:full_case_study3},~\ref{fig:full_case_study4},~\ref{fig:full_case_study5},and~\ref{fig:full_case_study6}.
By analyzing the T1 to T4 cases, we can draw the following conclusions about LLMs’ capabilities in historical research: (1) Lack of fine-grained knowledge: In T1 and T2 task, the model often provides correct ``general definitions'', but when faced with rubric-required specific details (e.g., the names of specific punishments, or exact time points in the evolution of certain official titles), it is prone to omissions or hallucinations. (2) Temporal confusion: In the T3 (historical reasoning) task, the model tends to confuse institutional differences across dynasties (e.g., Tang–Song vs. Ming–Qing). For example, when explaining the ``metropolitan examination'' (Huishi, \zh{会试}), some models fail to accurately distinguish that the term became standardized in the Ming–Qing, and instead make a loose analogy by mapping it to the Tang–Song ``provincial examination'' (Shengshi, \zh{省试}). (3) Differences in hard-constraint adherence: T4 (\celun{}) shows the greatest variation among models. This is the most difficult task because it requires the model to satisfy both format constraints (the baguwen structure) and negative constraints (name taboos). Some LLMs (e.g., Gemini-3-Pro) perform excellently on both taboos and structure, while others (e.g., Qwen3-235B) received a score of 0 for failing to follow the taboo or format requirements.

\begin{figure*}[t]
    \centering
    % \begin{subfigure}{\linewidth}
        \centering
        \includegraphics[width=\linewidth]{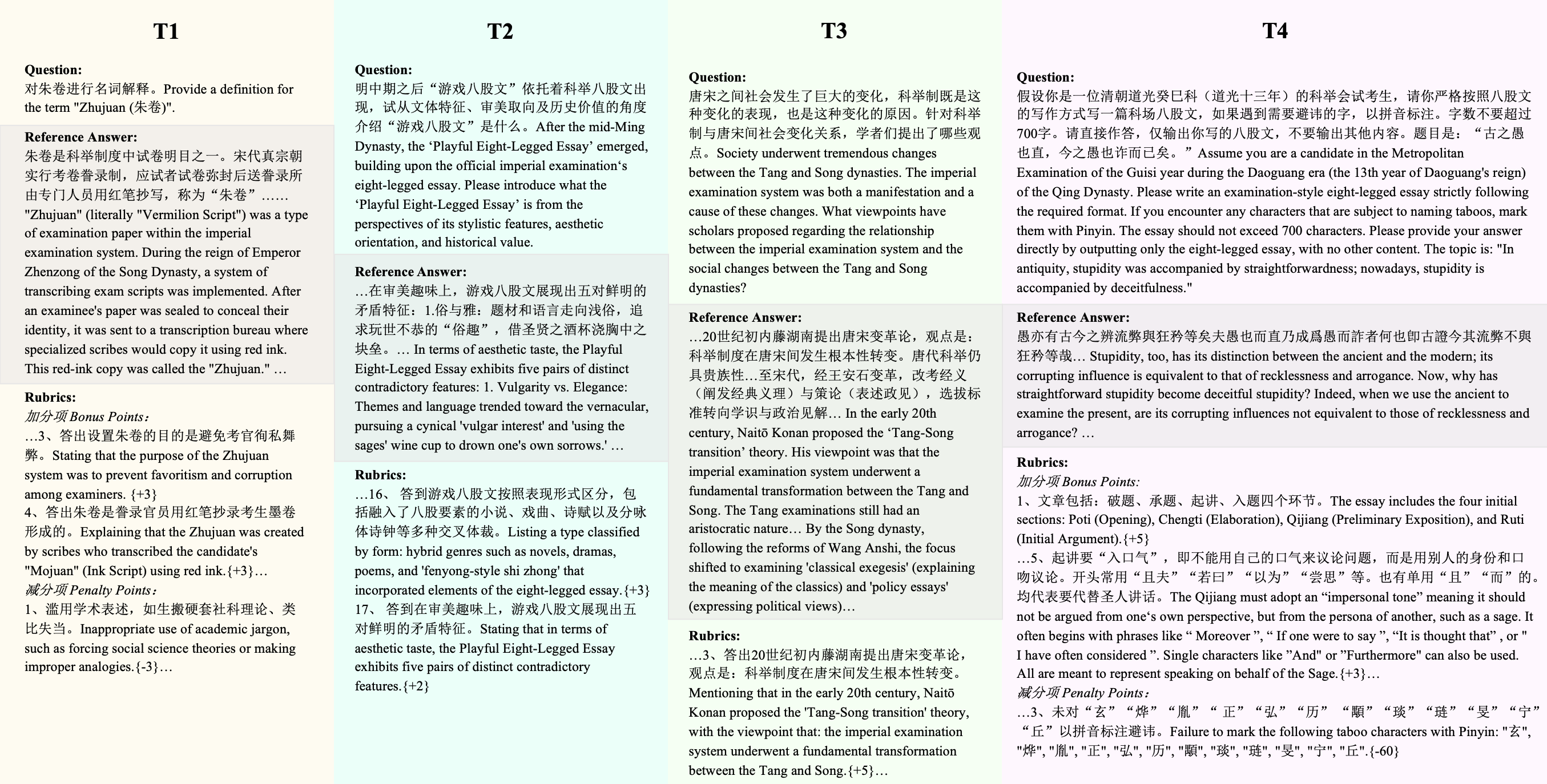}
    % \end{subfigure}
    \caption{Sample questions, reference answers, rubrics from four tasks.}
    \label{fig:sample_questions}
\end{figure*}

\section{Specific Source of ProHist-Bench}~\label{app-reference}
\textbf{Ancient Chinese Text}, including Ancient books, local gazetteers, original archival compilations, and eyewitness accounts:
\begin{itemize}
    \item \zh{（清）奎润等纂修，李兵、袁建辉点校：《钦定科场条例》，长沙：岳麓书社，2019年.}
    \item \zh{（清）素尔讷等纂修，霍有明、郭海文校注：《钦定学政全书校注》，武汉：武汉大学出版社，2009年.}
    \item \zh{《浙江通志》编纂委员会：《浙江通志》第76卷《教育志一》，杭州：浙江人民出版社，2019年.}
    \item \zh{顾廷龙主编：《清代朱卷集成》，台北：成文出版社，1992年.}
    \item \zh{李德龙、董玥：《未刊清代朱卷集成》，北京：学苑出版社，2019年.}
    \item \zh{陈维昭、张文达：《张文达藏稀见清代科举文献汇编》，桂林：广西大学出版社，2022年.}
    \item \zh{商衍鎏：《清代科举考试述录》，北京：故宫出版社，2012年.}
\end{itemize}

\textbf{Authoritative Scholarly Monographs} includes:
\begin{itemize}
\item \zh{齐如山：《中国的科名》，杭州：浙江古籍出版社，2020年.}
\item \zh{何怀宏：《选举社会及其终结——秦汉至晚清历史的一种社会学阐释》，北京：生活·读书·新知三联书店，1998年.}
\item \zh{李兵：《书院与科举关系研究》，武汉：华中师范大学出版社，2005年.}
\item \zh{李兵、刘海峰：《科举不只是考试》，上海：上海教育出版社，2018年.}
\item \zh{李华瑞主编：《“唐宋变革论”的由来与发展》，天津：天津古籍出版社，2010年.}
\item \zh{冯建民：《清代科举与经学关系研究》，武汉：华中师范大学出版社，2016年.}
\item \zh{贺晓燕：《清代科举落第制度研究》，广州：广东人民出版社，2022年.}
\item \zh{蒋维明：《李调元传》，北京：天地出版社，2024年.}
\item \zh{陈光新编著：《中国筵席宴会大典》，青岛：青岛出版社，1995年.}
\item \zh{翟国璋：《中国科举辞典》，南昌：江西教育出版社，2006年.}
\item \zh{夏征农：《辞海（中国古代史分册）》，上海：上海辞书出版社，1988年.}
\item \zh{杨金鼎主编，上海师范大学古籍整理研究所编：《中国文化史词典》，杭州：浙江古籍出版社，1987年.}
\item \zh{吴宗国：《唐代科举制度研究》，北京：北京大学出版社，2022年.}
\item \zh{高福顺：《辽朝科举制度》，吉林大学博士学位论文，2008年.}
\item \zh{申万里：《元代教育研究》，武汉：武汉大学出版社，2007年.}
\item \zh{姚大力：《蒙元制度与政治文化》，北京：北京大学出版社，2011年.}
\item \zh{梁庚尧：《北宋的改革与变法》，台北：东方出版中心，2024年.}
\item \zh{梁庚尧：《宋代科举社会》，台北：台湾大学出版中心，2015年.}
\item \zh{祝尚书：《宋代科举与文化》，北京：中华书局，2023年.}
\item \zh{陈宝良：《明代儒学生员与地方社会》，北京：中国社会科学出版社，2005年.}
\item \zh{郭培贵：《明代选举志考论》，北京：中华书局，2006年.}
\item \zh{龚笃清：《中国八股文史·明代卷》，长沙：岳麓书社，2017年.}
\item \zh{王颖、黄强：《游戏八股文研究》，武汉：武汉大学出版社，2015年.}
\item \zh{安东强：《清代学政规制与皇权体制》，北京：社会科学文献出版社，2017年.}
\item \zh{关晓红：《科举停废与近代中国》，北京：社会科学文献出版社，2017年.}
\item \zh{韩策：《科举改制与最后的进士》，北京：社会科学文献出版社，2017年.}
\item \zh{李细珠：《张之洞与清末新政研究》，北京：中国社会科学出版社，2015年.}
\item \zh{王建朗、黄克武主编：《两岸新编中国近代史·晚清卷》，北京：社会科学文献出版社，2016年.}
\item \zh{罗志田：《权势转移：近代中国的思想、社会与学术》，武汉：湖北人民出版社，1999年.}
\item \zh{王先明：《近代绅士——一个封建阶层的历史命运》，天津：天津人民出版社，1997年.}
\item \zh{费孝通：《中国绅士》，北京：中国社会科学出版社，2006年.}
\item \zh{费孝通：《乡土重建》，长沙：岳麓书社，2012年.}
\item \zh{张希清、毛佩琦、李世愉主编；张希清著：《中国科举制度通史（宋代卷）》，上海：上海人民出版社，2015年.}
\item \zh{张希清、毛佩琦、李世愉主编；金滢坤著：《中国科举制度通史（唐代卷）》，上海：上海人民出版社，2015年.}
\item \zh{张希清、毛佩琦、李世愉主编；李世愉、胡平著：《中国科举制度通史（清代卷）》，上海：上海人民出版社，2015年.}
\item \zh{张希清、毛佩琦、李世愉主编；郭培贵著：《中国科举制度通史（明代卷）》，上海：上海人民出版社，2015年.}
\item \zh{张希清、毛佩琦、李世愉主编；武玉环等著：《中国科举制度通史（辽金元卷）》，上海：上海人民出版社，2015年.}
\item \zh{乔卫平著，李国钧、王炳照主编：《中国教育制度通史（宋辽金元）》，济南：山东教育出版社，2000年.}
\item \zh{刘海峰：《科举学导论（增订本）》，北京：中国社会科学出版社，2025年.}
\item \zh{刘海峰：《科举学十讲》，杭州：浙江古籍出版社，2025年.}
\item \zh{王日根等著，刘海峰主编：《中国科举通史（清代卷）》，北京：人民出版社，2020年.}
\item \zh{金滢坤著，刘海峰主编：《中国科举通史（唐代卷）》，北京：人民出版社，2020年.}
\item \zh{钱建状著，刘海峰主编：《中国科举通史（宋代卷）》，北京：人民出版社，2020年.}
\item \zh{李兵著，刘海峰主编：《中国科举通史（辽金元卷）》，北京：人民出版社，2020年.}
\item \zh{郭培贵著，刘海峰主编：《中国科举通史（明代卷）》，北京：人民出版社，2020年.}
\item \zh{宫崎市定：《科举史》，郑州：大象出版社，2020年.}
\item \zh{李成茂：《高丽朝鲜两朝的科举制度》，北京：北京大学出版社，1993年.}
\item \zh{费正清、赖肖尔：《中国：传统与变革》，陈仲丹等译，南京：江苏人民出版社，2012年.}
\item \zh{Benjamin A. Elman：A Cultural History of Civil Examinations in Late Imperial China，Berkeley and Los Angeles：University of California Press，2000.}
\item \zh{Henrietta Harrison：The Man Awakened from Dreams: One Man’s Life in a North China Village, 1857–1942，Stanford：Stanford University Press，2005.}
\end{itemize}

\textbf{Top-tier Academic Papers} includes:
\begin{itemize}
\item \zh{安东强：《清末废八股后的四书义和五经义》，《文学遗产》，2015年第5期.}
\item \zh{安东强：《“中国政治史事论”与清末科举改制》，《文学遗产》，2021年第5期.}
\item \zh{安东强：《论校注本〈钦定学政全书〉的文献价值》，《武汉大学学报（人文科学版）》，2012年第1期.}
\item \zh{陈长文：《明代进士登科录的文献价值及其局限性》，《甘肃社会科学》，2006年第6期.}
\item \zh{陈长文：《明代科举中的官年现象》，《史学月刊》，2006年第11期.}
\item \zh{陈长文：《明代科举中的“告殿”现象》，《图书馆杂志》，2008年第4期.}
\item \zh{陈胜：《一项备受争议的教育制度——清末学堂奖励出身制度述评》，《华东师范大学学报（教育科学版）》，2011年第1期.}
\item \zh{陈文：《试析法国人对越南科举考试的影响》，载刘海峰、朱华山主编：《科举学的拓展与深化》，武汉：华中师范大学出版社，2013年.}
\item \zh{陈维昭：《论清代科场条例修订与八股文体演变的关系》，《文艺理论研究》，2025年第4期.}
\item \zh{陈宝良：《明代学官制度探析》，《社会科学辑刊》，1994年第3期.}
\item \zh{丁修真：《举人的路费：明代的科举、社会与国家》，《中国经济史研究》，2018年第1期.}
\item \zh{龚延明、高明扬：《清代科举八股文的衡文标准》，《中国社会科学》，2005年第4期.}
\item \zh{龚笃清：《试述明代前期八股文对文学的影响》，《中国文学研究》，2005年第1期.}
\item \zh{郭培贵：《关于明代科举研究中几个流行观点的商榷》，《清华大学学报（哲学社会科学版）》，2009年第6期.}
\item \zh{郭培贵：《明代庶吉士群体构成及其特点》，《历史研究》，2011年第6期.}
\item \zh{郭培贵：《明代武举的形成与确立》，《明史研究》，2017年.}
\item \zh{郭培贵：《明代科举中的座主、门生关系及其政治影响》，《中国史研究》，2012年第4期.}
\item \zh{郭培贵：《明代会试分卷录取制创立、实施及变迁考实》，《史学集刊》，2024年第6期.}
\item \zh{郭培贵：《试论明代提学制度的发展》，《文献》，1997年第4期.}
\item \zh{高福顺：《辽朝礼部贡院与知贡举考论》，《考试研究》，2011年第2期.}
\item \zh{关晓红：《晚清议改科举新探》，《史学月刊》，2007年第10期.}
\item \zh{关晓红：《清代朝考之创制与终结》，《学术研究》，2016年第11期.}
\item \zh{韩策：《科举改制与诏开进士馆的缘起》，《近代史研究》，2015年第1期.}
\item \zh{黄强：《游戏八股文的文学趣味——介绍俗文学的一个新品种》，《江南大学学报（人文社会科学版）》，2011年第1期.}
\item \zh{贾安琪、聂鑫：《清代〈科场条例〉纂修考述》，《浙江大学学报（人文社会科学版）》，2024年第11期.}
\item \zh{姜新：《评清末民初的留学生归国考试》，《史学月刊》，2005年第12期.}
\item \zh{李发根：《科举制的废除与近代中国乡村危机研究》，《山西师范大学学报（哲学社会科学版）》，2016年第6期.}
\item \zh{李建军：《明代武举制度述略》，《南开学报》，1997年第3期.}
\item \zh{李思成：《明清会试“阅本经”规则的演变》，《山东社会科学》，2023年第9期.}
\item \zh{李思成：《身不由己：明代的座主门生与党争再探》，《山东社会科学》，2025年第1期.}
\item \zh{李永明：《京师大学堂历史学科的发生与发展》，《史学史研究》，2025年第3期.}
\item \zh{林浩彬：《裁撤府、州、县学教职与清末新政》，《社会科学战线》，2025年第9期.}
\item \zh{刘海峰：《“科举”含义与科举制的起始年份》，《厦门大学学报（哲学社会科学版）》，2008年第5期.}
\item \zh{刘海峰、毛鹏程：《清乾隆朝科举改革：动因、举措及影响》，《厦门大学学报（哲学社会科学版）》，2023年第2期.}
\item \zh{刘海峰：《论述东亚科举文化圈的形成与演变》，《厦门大学学报（哲学社会科学版）》，2016年第5期.}
\item \zh{刘海峰：《中国对日韩越三国科举的影响》，《学术月刊》，2006年第12期.}
\item \zh{刘明鑫：《明代的科举走报》，《史学月刊》，2019年第7期.}
\item \zh{刘明鑫：《明代会试考生路费资助制度考论》，《历史档案》，2020年第4期.}
\item \zh{刘明鑫：《明代会试考生应考旅费考察》，《中国史研究》，2022年第4期.}
\item \zh{刘晓琴：《严复与晚清留学生归国考试研究》，《南开学报》，2014年第1期.}
\item \zh{刘希伟：《清代科举考试中的“商籍”考论——一种制度史的视野》，《清史研究》，2010年第3期.}
\item \zh{刘志强：《越南阮朝科举及其本土特色》，《东南亚纵横》，2010年第4期.}
\item \zh{兰婷、王成铭：《金代女真官学》，《社会科学战线》，2010年第9期.}
\item \zh{毛晓阳：《清代宾兴礼考述》，《清史研究》，2007年第3期.}
\item \zh{宋豪飞：《清代科举教科书：<钦定四书文>的编选与文风宗尚》，《中山大学学报（社会科学版）》，2025年第3期.}
\item \zh{宋巧燕：《清代科举试帖诗写作规范探析》，《教育与考试》，2015年第3期.}
\item \zh{申万里、窦相国：《元代儒户体制下士人的管理制度》，《社会科学研究》，2025年第5期.}
\item \zh{薛瑞兆：《论金朝的“南北选”》，《学术交流》，2025年第1期.}
\item \zh{郗志群：《封建科举、职官中的“官年”——从杨守敬的乡试硃卷谈起》，《历史研究》，2003年第4期.}
\item \zh{萧启庆：《元代的儒户：儒士地位演进史上的一章》，载《元代史新探》，台北：新文丰出版公司，1983年.}
\item \zh{萧启庆：《元代科举特色新论》，《台北“中研院”历史语言研究所集刊》，第81本，2010年.}
\item \zh{王静：《清代科举中“岁科连考”的规制变通与实践效果》，《清史研究》，2025年第4期.}
\item \zh{王庆成：《清代学政官制之变化》，《清史研究》，2008年第1期.}
\item \zh{王日根、章广：《清代八旗科举制度的发展及其影响》，《考试研究》，2015年第5期.}
\item \zh{王熹：《明代朝野对科举制度的评论》，《明史研究》，2001年.}
\item \zh{王学深：《清代科举试卷违式问题探析》，《古代文明（中英文）》，2024年第4期.}
\item \zh{王学深：《清代乾隆朝科举冒籍问题论述》，《中国考试》，2016年第4期.}
\item \zh{王学深：《清代科举“未殿试”成因再探析》，《地域文化研究》，2022年第4期.}
\item \zh{吴恩荣：《科考、遗才与大收：明代乡试资格考试述论》，《安徽大学学报（哲学社会科学版）》，2013年第5期.}
\item \zh{吴光辉：《科举考试与日本》，《东南学术》，2005年第4期.}
\item \zh{王方：《科举制度与儒学在日本的早期传播》，《外文研究》，2004年第2期.}
\item \zh{杨胜祥：《清代科举“补殿试”对进士名次的影响》，《清史研究》，2020年第2期.}
\item \zh{于晓燕：《“义学”释义》，《贵州师范学院学报》，2014年第10期.}
\item \zh{张仲民：《“科举”余绪：清末最后的“杂试”与“朝考”》，《史学月刊》，2024年第7期.}
\item \zh{张晓波：《朝鲜王朝科举制度研究》，山东师范大学博士学位论文，2020年.}
\item \zh{郑天挺：《清代考试的文字——八股文和试帖诗》，《故宫博物院院刊》，1982年第2期.}
\item \zh{赵利峰：《清中后期广东闱姓考原》，《暨南史学》，2003年.}
\item \zh{左玉河：《论清季学堂奖励出身制》，《近代史研究》，2008年第4期.}
\end{itemize}
\end{document}